\documentclass[runningheads]{llncs}

\usepackage[mobile]{eccv}

\usepackage{eccvabbrv}

\usepackage[accsupp]{axessibility}

\usepackage{algorithm}
\usepackage{algpseudocode}
\usepackage{amsfonts}
\usepackage{amsmath}
\usepackage{amssymb}
\usepackage{bm}
\usepackage{mathtools}

\usepackage{array}
\usepackage{booktabs}
\usepackage{graphicx}
\usepackage{makecell}
\usepackage{multirow}
\usepackage{tabularx}
\usepackage{microtype}
\usepackage[section]{placeins} %
\usepackage{caption} %

\usepackage{listings}
\usepackage[most]{tcolorbox}
\usepackage{xcolor}
\definecolor{cvprblue}{rgb}{0.21,0.49,0.74}
\definecolor{darkgreen}{HTML}{006400}

\usepackage{url}
\usepackage{xspace}

\usepackage{silence}
\usepackage[numbers,sort&compress]{natbib}
\renewcommand{\refname}{References}

\makeatletter
\@ifundefined{assumption}{}{}
\makeatother

\def\ours{VIGA\xspace}

\newtcolorbox{promptbox}{
  colback=white,      
  colframe=black,     
  boxrule=0.8pt,      
  arc=0mm,            
  left=4pt,right=4pt, 
  top=4pt,bottom=4pt,
  breakable           
}

\lstdefinelanguage{json}{
    basicstyle=\ttfamily\small,
    numbers=left,
    numberstyle=\tiny\color{gray},
    keywordstyle=\color{blue},
    stringstyle=\color{teal},
    commentstyle=\color{gray},
    breaklines=true,
    literate=
     *{0}{{{\color{blue}0}}}{1}
      {1}{{{\color{blue}1}}}{1}
      {2}{{{\color{blue}2}}}{1}
      {3}{{{\color{blue}3}}}{1}
      {4}{{{\color{blue}4}}}{1}
      {5}{{{\color{blue}5}}}{1}
      {6}{{{\color{blue}6}}}{1}
      {7}{{{\color{blue}7}}}{1}
      {8}{{{\color{blue}8}}}{1}
      {9}{{{\color{blue}9}}}{1}
}

\usepackage[pagebackref,breaklinks,colorlinks,citecolor=eccvblue]{hyperref}

\usepackage{orcidlink}

\begin{document}

\title{Vision-as-Inverse-Graphics Agent \\ via Interleaved Multimodal Reasoning}

\author{
    Shaofeng Yin\inst{1,4}$^\ast$ \and
    Jiaxin Ge\inst{1} \and
    Zora Zhiruo Wang\inst{2} \and
    Chenyang Wang\inst{1,4} \and
    Xiuyu Li\inst{1} \\
    Michael J. Black\inst{3} \and
    Trevor Darrell\inst{1} \and
    Angjoo Kanazawa\inst{1} \and
    Haiwen Feng\inst{1,3,4}$^\dagger$
}

\institute{
    $^1$University of California, Berkeley \qquad
    $^2$Carnegie Mellon University \\
    $^3$Max Planck Institute for Intelligent Systems \qquad
    $^4$Impossible, Inc. \\
    \vspace{1.5mm}
    $^\ast$\email{shaofeng.yin@impossible.place} \qquad
    $^\dagger$\email{haiwen.feng@berkeley.edu} \\
    \vspace{2mm}
    {
        \ttfamily \small \centering
        \href{https://fugtemypt123.github.io/VIGA-website/}{
            \raisebox{-0.3em}{\includegraphics[height=1.5em]{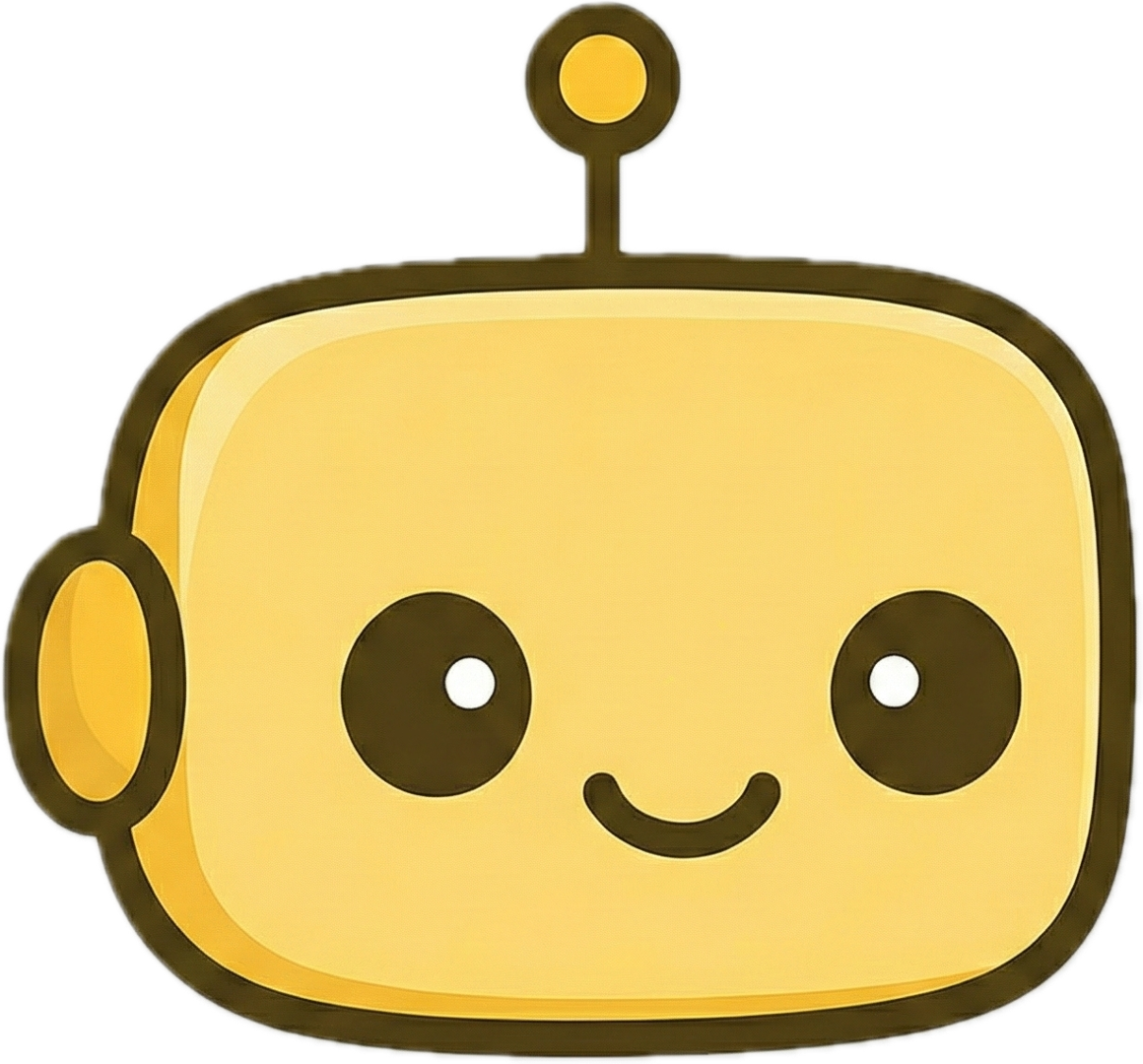}}
            fugtemypt123.github.io/VIGA-website
        }
    }
}

\authorrunning{Shaofeng Yin et al.}

\maketitle

\par 
\begingroup
\nolinenumbers 
\noindent\centering
\begin{minipage}{0.85\textwidth}
    \centering
    \includegraphics[width=\linewidth]{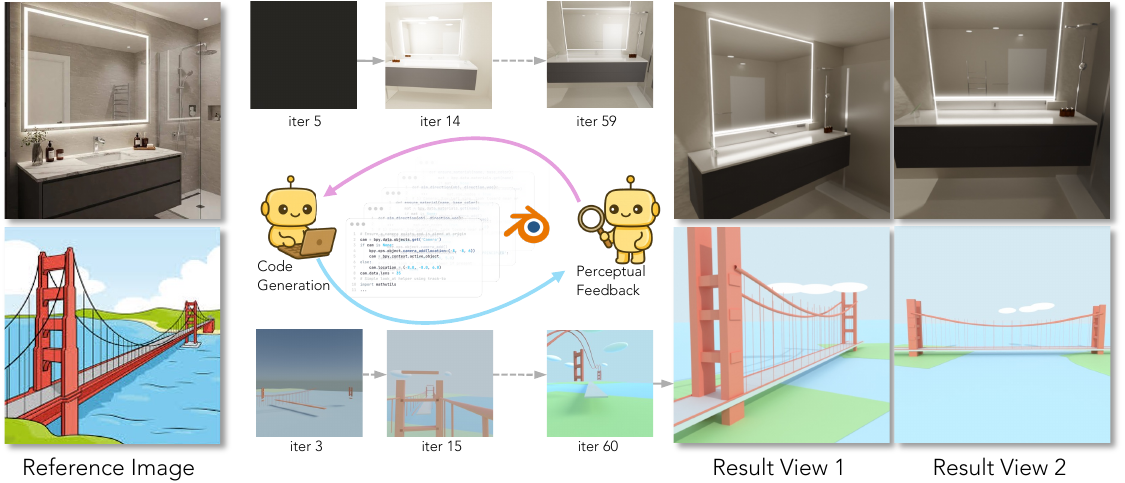}
    \captionsetup{font=small,hypcap=false}
    \captionof{figure}{\textbf{\ours{} constructs 3D scenes as executable programs from a reference image.} 
    Starting from an empty 3D environment, \ours{} constructs the scene by leveraging abstract \textit{symbolic generation} (writing and executing code to modify the scene) and active \textit{visual verification} (rendering the state to inspect visual discrepancies and provide grounded feedback). By seamlessly interleaving code execution with fine-grained visual perception, the agent progressively refines the layout, geometry, and lighting to match the reference image.}
    \label{fig:teaser_multi_agent} 
\end{minipage}
\par 
\endgroup

\begin{abstract}

Vision-as-inverse-graphics, the concept of reconstructing images into editable programs, remains challenging for Vision-Language Models (VLMs), which inherently lack fine-grained spatial grounding in one-shot settings. 
To address this, we introduce \ours{} (Vision-as-Inverse-Graphics Agent), an interleaved multimodal reasoning framework where symbolic logic and visual perception actively cross-verify each other. \ours{} operates through a tightly coupled \textit{code-render-inspect} loop: synthesizing symbolic programs, projecting them into visual states, and inspecting discrepancies to guide iterative edits. 
Equipped with high-level semantic skills and an evolving multimodal memory, \ours{} sustains evidence-based modifications over long horizons. 
This training-free, task-agnostic framework seamlessly supports 2D document generation, 3D reconstruction, multi-step 3D editing, and 4D physical interaction. Finally, we introduce BlenderBench, a challenging visual-to-code benchmark. Empirically, \ours{} substantially improves accuracy compared with one-shot baselines in BlenderGym (\textit{35.32\%}), SlideBench (\textit{117.17\%}) and our proposed BlenderBench (\textit{124.70\%}).
\end{abstract}

\section{Introduction}

Consider the images in~\cref{fig:teaser_multi_agent}. What would it take to recreate these scenes symbolically in a graphics engine like Blender? Starting from scratch, we would need to write code that crafts or imports appropriate assets, establishes scene layout, places objects, and configures materials and lighting. Such an ability would yield compositional scene representations, easily manipulated for applications like robot training or interpretable digital twins. This capability represents achieving ``vision-as-inverse-graphics'', a longstanding aspiration since Larry Roberts' seminal 1963 thesis on the blocks world \citep{Roberts1963_MachinePerception3D}. 

However, solving inverse graphics for complex scenes \citep{nsd, Yi2018_NeuralSymbolicVQA} remains extremely difficult. Accurate reconstruction is not a passive, one-shot generation problem; it inherently requires continuous observation, active information-seeking, and precise geometric refinement.
Existing approaches largely fall into two paradigms. 
On the one hand, differentiable inverse graphics~\citep{Blanz1999AMM, Loper2014OpenDRAA, Liu2019SoftRA, Feng:SIGGRAPH:2021} provides fine-grained visual grounding through pixel-level optimization. 
However, its strict reliance on local gradients makes it vulnerable to local minima, especially when the discrepancy is large between the initial state and the target reference. Fundamentally lacking the capacity for high-level semantic reasoning and active exploration, the optimization process is unable to leap out of these non-convex traps. 
On the other hand, Vision-Language Models (VLMs) offer a complementary paradigm. Leveraging broad semantic priors, VLMs can generate scene code in a single pass, bypassing the local minima that trap gradient-based methods. However, this one-shot paradigm is inherently open-loop. 
Without fine-grained visual grounding, they struggle to precisely verify and refine complex scene details, such as camera extrinsics, object poses, and lighting. Consequently, even powerful proprietary VLMs struggle with one-shot code generation for accurate complex scene reconstruction \citep{sun20253dgeneralistselfimprovingvisionlanguageactionmodels}.

In this work, we demonstrate that these limitations can be overcome by leveraging their complementary strengths. We present \ours{} (Vision-as-Inverse-Graphics Agent), a multimodal coding agent that tightly couples discrete symbolic generation with continuous visual perception. By operating through a concise \textit{code-render-inspect} loop, \ours{} allows code logic and visual evidence to actively cross-verify each other. From a modern agentic perspective, this is a direct instantiation of interleaved multimodal reasoning. It aligns with the emerging notion of ``thinking with images'', but in a substantially richer setting: rather than relying on simple 2D image operations (\eg, crop/zoom), \ours{} reasons through a graphics engine whose action space is comprehensive, including camera poses, 3D geometry, materials, lighting, visibility, timestamps, etc. In a nutshell, this agent thinks with a graphics engine.

\begin{figure*}[t]
    \centering
    \includegraphics[width=\linewidth]{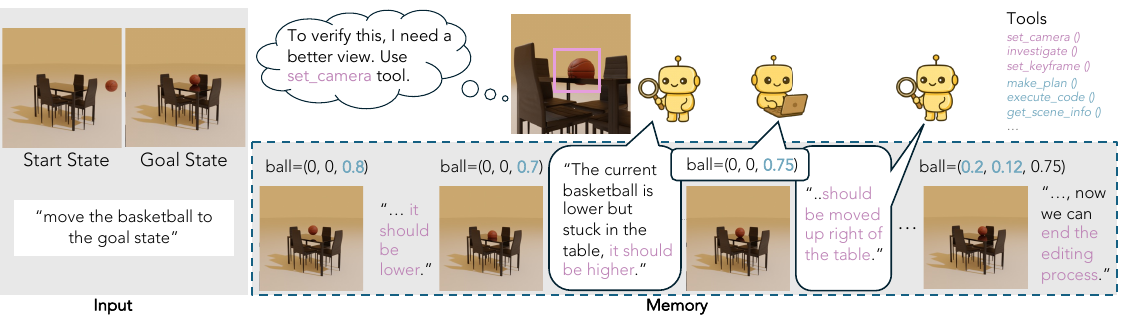}
    \caption{\textbf{The main pipeline of \ours{}.} \ours{} operates through a continuous \textit{code-render-inspect} loop. At each step, the agent synthesizes a program, which is executed to render a new scene. The agent then actively inspects this scene by invoking read-only perceptual interfaces to adjust camera viewpoints, identify the dominant discrepancy, and feed this visual feedback into the next step. We illustrate the scene editing task here. \ours{} can also synthesize scenes from scratch given a goal specification (as in~\cref{fig:teaser_multi_agent}).}
    \label{fig:framework}
\end{figure*}

To robustly execute this multi-turn reasoning (\cref{fig:framework}), \ours{} relies on two critical pillars. First, it equips the agent with a \textbf{high-level semantic skill library} that strictly decouples scene observation from modification. For instance, the agent can repeatedly invoke read-only perceptual interfaces (\eg, \texttt{investigate}) to actively shift camera viewpoints and gather multi-angle visual evidence without altering the underlying scene. By explicitly comparing these rendered multi-angle views against the target image to localize discrepancies, the agent then triggers execution interfaces (\eg, \texttt{execute\_code}) to apply targeted programmatic edits. Second, it leverages an \textbf{evolving, sliding-window multimodal context memory} to execute evidence-based modifications. This structured memory retains recent plans, code diffs, and render histories. Crucially, its sliding-window mechanism actively bounds this retained history, inherently preventing context bloat and rot across long-horizon trajectories.

\ours{} is a zero-shot, training-free framework. The verification process relies only on the agent's ability to interpret discrepancies between renders and reference images. Therefore it requires no auxiliary predictors or differentiable renderers. This design makes the approach exceptionally task and model-agnostic. It seamlessly generalizes across 2D document layout editing (\cref{fig:demo_2d}), 3D scene reconstruction (\cref{fig:demo_3d}), multi-step scene editing (\cref{fig:trajectory_tasks}), and 4D physical interaction (\cref{fig:demo_4d}). Under a unified protocol, \ours{} serves as a powerful capability testbed to evaluate heterogeneous foundation VLMs. To rigorously stress-test this unified paradigm, we introduce \textbf{BlenderBench}, a challenging suite of 27 tasks covering spatial adjustments, progressive editing, and compositional generation. \ours{} delivers substantial gains across diverse benchmarks, yielding signficant improvements on BlenderGym (\textit{+35.32\%}), SlideBench (\textit{+117.17\%}), and BlenderBench (\textit{+124.70\%}).

In summary, we make the following contributions: 
(1) We introduce \ours{}, a training-free, execution-grounded agent that reconstructs and edits scenes by closing the loop between code synthesis, rendering, and visual verification. This reframes vision-as-inverse-graphics as a coding agent problem, where the rendered world becomes a medium for interleaved multimodal reasoning. 
(2) We propose a robust architectural paradigm combining a high-level semantic skill library with an evolving multimodal context memory, effectively sustaining long-horizon cross-modal reasoning without finetuning. 
(3) We demonstrate that \ours{} delivers substantial empirical gains across a diverse spectrum of tasks, seamlessly supporting 2D document layout, 3D reconstruction, multi-step scene editing, and 4D physical interaction. 
(4) We release BlenderBench, a challenging, open-ended benchmark designed to rigorously stress-test agentic inverse graphics capabilities, where \ours{} shows significant improvements against existing baselines.

\section{Related Work}
\label{sec:related_work}
\noindent\textbf{LLM-based Agent.}
LLM-based agents that observe and act in digital environments, such as browsers \citep{yao2022webshop} and computers \citep{xie2024osworld}, have achieved impressive progress across various tasks, such as resolving issues in software engineering~\citep{yang2024swe, he2025llmswe, tawosi2025almasautonomousllmbasedmultiagent}, or navigating through web pages to seek information or complete tasks on behalf of the users~\citep{zhou2023webarena, qin2025ui, azam2025reflectionbasedmemorywebnavigation}.
However, most agents behave effectively only when perceiving the environment as text and fall short when processing visual information, such as selecting products of certain styles on shopping websites \citep{koh2024visualwebarena}. These text-centric agents become even more limited when situated in visual environments such as 2D or 3D spaces \citep{gupta2023visual,suris2023vipergpt, Chen_2024_CVPR, Hong2023_3DLLM, wang2024spatial, cheng2024spatialrgpt, Dwedari2023_3DSceneVQA, Zhang2023_ModularEmbodiedAgents, sun20253d, Kodnongbua2023_ReparamCAD, Xue2023_ULIP}, thus often fall short in producing high-quality visuals \citep{ge2025autopresent, gu2025blendergym}.
In this work, we bridge this gap by presenting a unified agent that performs interleaved multimodal reasoning to iteratively synthesize and analyze visuals across 2D, 3D, and 4D spaces.
We facilitate multi-turn verification and refinement through adaptive agent memory \citep{wang2025agent,suzgun2025dynamic,zhang2025agentic, xu2025amem, ye2025taskmemoryenginespatial}.

\noindent\textbf{Visual Analysis or Synthesis with Code.}
Structured code has been used separately for image analysis and synthesis.
On the analysis side, visual programming approaches~\citep{gupta2023visual, suris2023vipergpt, subramanian2023modular, ge2024recursive, hu2023visual} generate code to solve visual understanding tasks such as VQA.
However, these approaches generate code in a single pass and do not edit the code based on any visual feedback.
On the synthesis side,
procedural content generation approaches~\citep{infinigen2023infinite, infinigen2024indoors, procthor} rely on predefined rules to generate specific 3D content.
Recently, instruction-driven approaches generate visual content from text instructions leveraging large language models, spanning SVG generation~\citep{yang2025omnisvg, Wu2024Chat2SVGVG, xing2024llm4svg, wang2025svgen},
slide generation~\citep{ge2025autopresent, zheng2025pptagent, bandyopadhyay-etal-2024-enhancing-presentation, tang2025slidecoderlayoutawareragenhancedhierarchical},
and 3D scene generation~\citep{sun20253d, hu2024scenecraft, sun20253dgeneralistselfimprovingvisionlanguageactionmodels, ling2025scenethesis, layoutgpt2023, sun2025layoutvlm, aguinakang2024openuniverseindoorscenegeneration, zhou2025scenex, Qin2025ApplyHT, sceneteller2024}.
These approaches lack effective analysis that can ground the scene back to code, and thus fail to update the scene with code editing.
We unify the two lines of work within a single framework of interleaved multimodal reasoning that executes programs, verifies generation, and then edits code in an iterative analysis-by-synthesis loop.

\noindent\textbf{Vision-as-Inverse-Graphics through Programs.}
Since Larry Roberts’s seminal Blocks World thesis \citep{Roberts1963_MachinePerception3D}, computer vision has often been framed as the inverse of computer graphics. A stricter interpretation of this perspective aims to recover a \emph{graphics program} \citep{progsyn2017}, \ie, an interpretable, compositional abstraction of a scene, from natural images using neural–symbolic methods \citep{neurosymbol2023}. For instance, PICTURE \citep{Kulkarni2015_PICTURE} introduced a probabilistic programming framework for representing arbitrary 2D and 3D scenes, demonstrating early successes on faces, bodies, and objects.
Similarly, Wu~\etal\citep{nsd} proposed inferring structured markup code from images, which can be rendered directly. Although effective for multi-object abstract scenes like clip-art or Minecraft, their approach struggled to generalize to real images.
More recent approaches leverage large language models (LLMs) to generate scene programs and refine them via supervision or verification feedback \citep{kulits2024re,gu2025blendergym,ling2025scenethesis,sun20253d}. While closely related to our work, these methods differ in two key respects.
First, they rely on synthetic datasets or domain-specific verification pipelines, which restrict scalability and adaptability to diverse forms of visual feedback. Second, their workflows are typically predefined or tailored to particular downstream tasks, such as 3D scene reconstruction using pre-defined graphics engines and model weights.

In contrast, \ours{} introduces a unified, training-free agent that seamlessly interleaves discrete symbolic code generation with active visual verification. While existing methods often rely on rigid, task-specific pipelines, our framework equips the agent with a versatile semantic skill library. The agent dynamically invokes these tools to probe the environment and execute changes, rather than being constrained by embedded, domain-specific architectures. By operating through an execution-grounded \textit{code-render-inspect} loop sustained by an evolving multimodal context memory, \ours{} inherently functions as a task-agnostic framework. This enables the agent to adapt zero-shot to a wide range of complex, long-horizon inverse graphics tasks---spanning from compositional 3D scene editing to dynamic 4D physical simulations---without requiring any model finetuning.

\section{Method}
\label{sec:3:method}

\begin{figure}[t]
    \centering
    \includegraphics[width=1\linewidth]{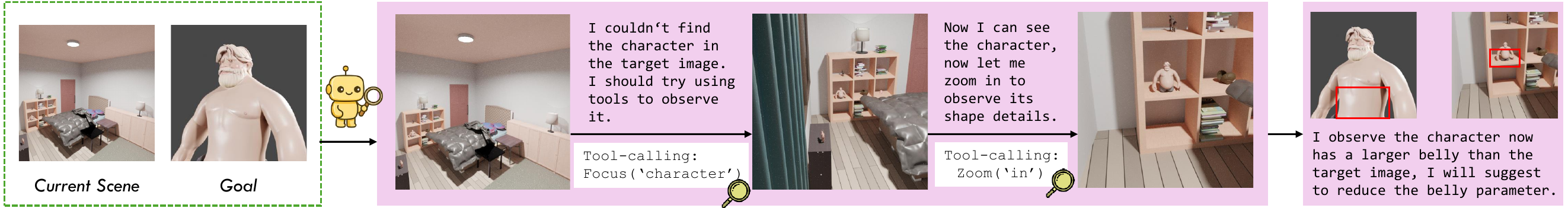}
    \caption{\textbf{Agentic Visual Navigation.} Demonstration of a visual inspection trajectory in a cluttered scene. The agent autonomously invokes spatial tools (focusing, zooming) to locate the inconspicuous target before evaluating attributes, showcasing an interleaved code-visual reasoning loop.}
    \label{fig:trajectory_navigation}
\end{figure}

\begin{figure}[t]
    \centering
    \includegraphics[width=1\linewidth]{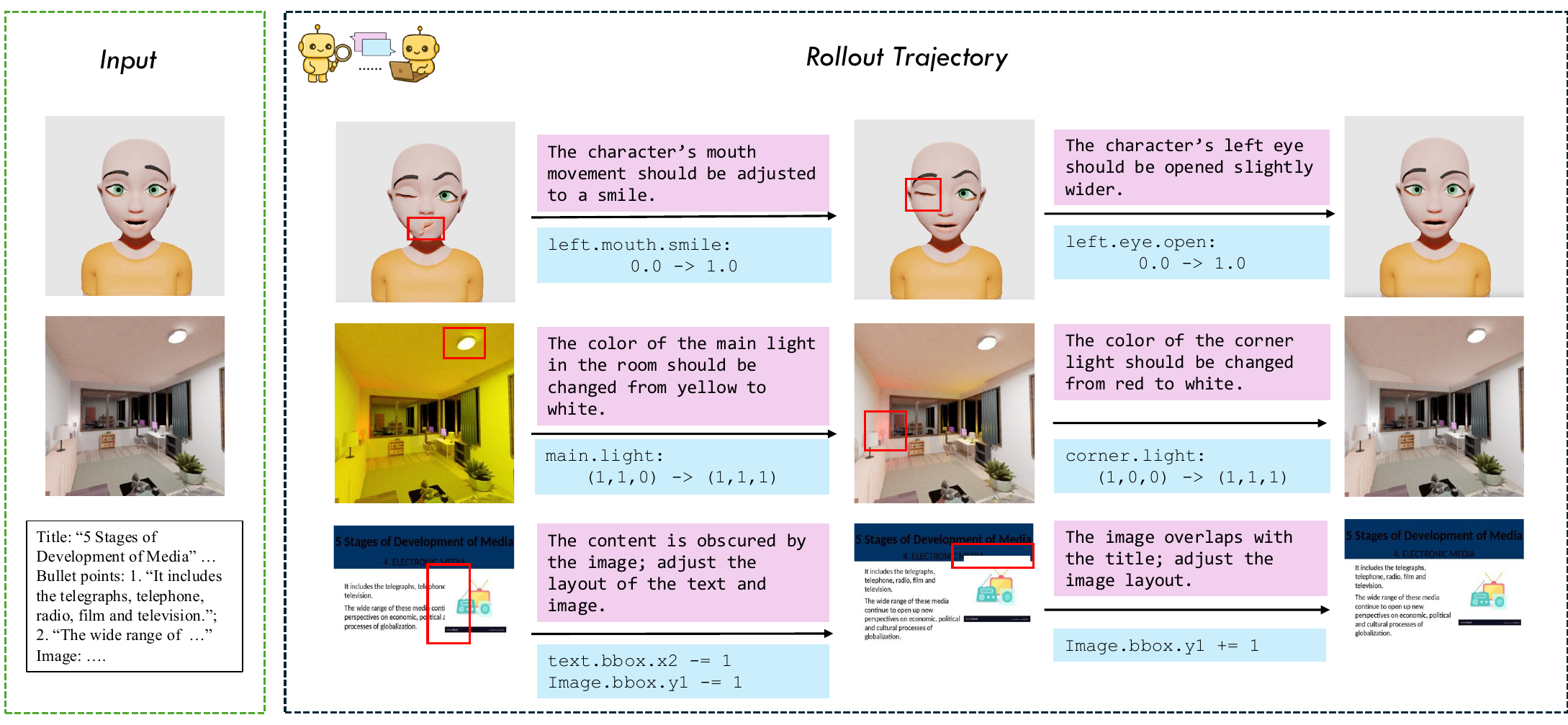}
    \caption{\textbf{Fine-Grained Visual Grounding.} Agent trajectories across different tasks. The agent detects nuanced visual discrepancies (\eg, mouth shape, lighting color) and dynamically maps these high-level observations directly to precise code parameters, avoiding rigid rule-based heuristics.}
    \label{fig:trajectory_tasks}
\end{figure}

Recognizing the complementary strengths of VLMs (for holistic semantic reasoning) and differentiable inverse graphics (for fine-grained visual refinement), we introduce \ours{}, an interleaved multimodal reasoning framework designed to leverage both paradigms. Rather than treating reasoning steps in isolation, \ours{} tightly couples these processes: the agent employs active visual navigation to systematically explore and overcome severe spatial misalignments (\cref{fig:trajectory_navigation}), while translating these fine-grained visual observations into targeted, programmatic code edits (\cref{fig:trajectory_tasks}). As illustrated in~\cref{fig:framework}, \ours{} operationalizes this continuous cross-modal loop via three core components. First, the agent drives an interactive reasoning paradigm across the visual and symbolic spaces (\cref{sec:3.1:reasoning}). To sustain this multi-turn trajectory, it leverages a dynamically evolving context memory (\cref{sec:3.2:memory}). Finally, it utilizes a high-level semantic skill library to reliably ground its perception and execution capabilities (\cref{sec:3.3:skills}). 

\begin{algorithm}[t]
\caption{\ours{} Interleaved Multimodal Reasoning Loop.}
\label{alg:abs-framework}
\begin{algorithmic}[1]
\Require Target specification $s_g^*$
\State Initialize contextual memory $M_0 \gets \emptyset$
\For{$t = 0,1,\dots,T-1$}
    \State \textcolor{gray}{// Phase 1: Symbolic Reasoning from Visual Input}
    \State $\tau_t^{plan} \gets \mathcal{A}(s_g^*, M_t)$ 
    \Comment{Formulates high-level plan}
    \If{$\tau_t^{plan}$ includes \texttt{end\_process}} \textbf{break} \EndIf
    \State $p_t \gets \mathcal{A}(s_g^*, M_t, \tau_t^{plan})$
    \Comment{Synthesizes executable program}
    
    \State \textcolor{gray}{// Phase 2: Cross-Modal Execution}
    \State $s_t \gets \mathrm{exec}(p_t)$
    \Comment{Renders concrete geometric states}
    
    \State \textcolor{gray}{// Phase 3: Visual-to-Symbolic Feedback}
    \State $\tau_t^{obs} \gets \mathcal{A}(s_g^*, s_t, M_t)$
    \Comment{Explores the rendered scene}
    
    \State $f_t \gets \mathcal{A}(s_g^*, s_t, \tau_t^{obs}, M_t)$
    \Comment{Translates into actionable feedback}
    
    \State \textcolor{gray}{// Phase 4: Context Memory Update (\cref{sec:3.2:memory})}
    \State $M_{t+1} \gets \mathrm{Tail}_L(M_t \cup \{p_t, f_t\})$
\EndFor
\end{algorithmic}
\end{algorithm}

\subsection{Interleaved Multimodal Reasoning Framework}

\label{sec:3.1:reasoning}

Formally, given an input target $s_g^*$ (\eg, a reference image) and an empty initial state $s_0$, the objective of inverse graphics is to synthesize a final executable program $p_T$ that renders a state $s_T$ visually consistent with $s_g^*$. We implement this iterative analysis-by-synthesis process via interleaved multimodal reasoning. Specifically, we formulate this as a dynamic execution loop that fluidly transitions across different multimodal action spaces, utilizing each modality to cross-verify and inherently ground the others. Rather than naively chaining isolated single-modal steps, our approach deeply integrates these spaces: the agent leverages active visual navigation to verify spatial configurations and expose discrepancies (\cref{fig:trajectory_navigation}), which subsequently guide precise programmatic edits in the symbolic space (\cref{fig:trajectory_tasks}), forming a robust, self-correcting reasoning cycle.

\noindent \textbf{Symbolic Reasoning from Visual Input.} 
At each iteration $t$, the agent $\mathcal{A}$ drives the generation loop forward. Conditioned on its current context memory $M_t$, it evaluates the target $s_g^*$ to derive a strategic planning trajectory $\tau_t^{plan} = \mathcal{A}(s_g^*, M_t)$. To instantiate this high-level plan, the agent must operate within the discrete symbolic space to synthesize exact programmatic actions. During this phase, $\mathcal{A}$ leverages forward anticipation—mentally mapping abstract symbolic parameters to their continuous visual consequences—to construct the valid code:
\begin{equation}
    p_t = \mathcal{A}(s_g^*, M_t, \tau_t^{plan}).
    \label{eq:action_planning}
\end{equation}

\noindent \textbf{Cross-Modal Execution.} 
Unlike standard coding tasks where the generated program $p_t$ serves as a terminal textual artifact, our framework utilizes $p_t$ to drive a deterministic modality shift. Instead of treating the code as an open-loop endpoint, we execute it within a 3D graphics engine to render a continuous visual state: $s_t = \mathrm{exec}(p_t)$. This execution acts as the crucial cross-modal bridge. By projecting abstract symbolic structure into concrete visual observations, this mapping structurally defines the interaction loop: the symbolic syntax provides a precise action space for modifying the environment, while the newly rendered visual state immediately exposes spatial discrepancies, directly driving the next iteration of visual-to-code feedback.

\noindent \textbf{Visual-to-Symbolic Feedback.} 
To close the iterative loop, the agent seamlessly transitions into an active observation phase within the continuous visual space. By employing interactive tools (\eg, adjusting camera viewpoints) to explore the newly rendered scene $s_t$, the agent $\mathcal{A}$ generates a comprehensive observation trajectory: $\tau_t^{obs} = \mathcal{A}(s_g^*, s_t, M_t)$. Finally, it rigorously compares this rendered outcome against the target specification $s_g^*$, translating any observed visual discrepancies into actionable natural language feedback. This feedback serves as the critical bridge back to the symbolic domain, guiding the agent's next iteration of programmatic edits:
\begin{equation}
    f_t = \mathcal{A}(s_g^*, s_t, \tau_t^{obs}, M_t).
    \label{eq:feedback}
\end{equation}
This grounded visual feedback $f_t$, alongside the synthesized program $p_t$, is subsequently assimilated to dynamically evolve the context memory into $M_{t+1}$, providing the critical foundation for the ensuing reasoning cycle detailed next.

\begin{figure}[t]
    \centering
    \includegraphics[width=\linewidth]{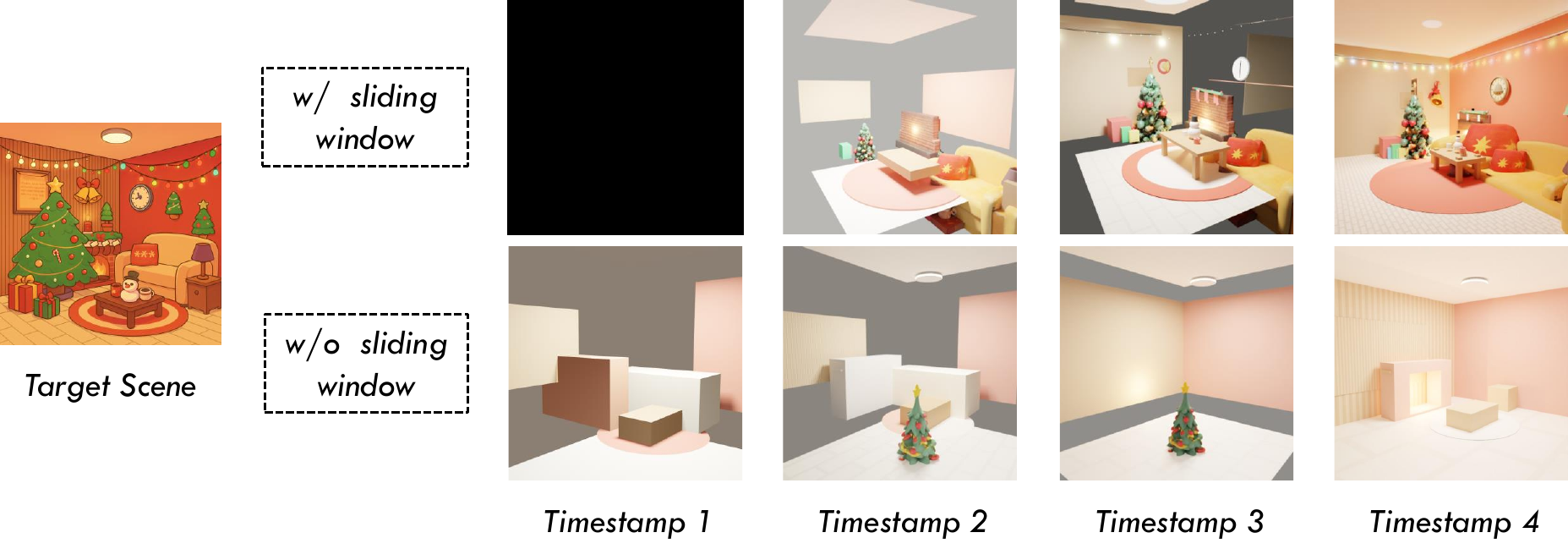}
    \caption{\textbf{Ablation on Evolving Multimodal Memory.} We compare the sequential generation process with (w/) and without (w/o) our sliding window memory mechanism. With the memory window, the agent successfully maintains long-horizon context, progressively building a coherent scene (tree, sofa, fireplace). Without it, the agent suffers from severe context loss, losing previously generated objects (\eg, tree) and spatial relationships in later iterations.}
    \label{fig:memory_ablation}
\end{figure}

\subsection{Evolving Multimodal Memory for Long-Horizon Reasoning}
\label{sec:3.2:memory}

Unlike standard text-based chat logs, the memory $M_t$ is constructed as a structured multimodal trajectory. At each iteration, it systematically archives the interleaved sequence of reasoning traces (plans and feedback $f_t$), symbolic actions (executable programs $p_t$), and grounded observations (rendered states $s_t$). This structured formulation is indispensable for fine-grained visual grounding. As illustrated in \cref{fig:framework}, maintaining this multimodal historical context empowers the agent to perform complex, progressive spatial refinements. By explicitly referencing past programmatic attempts alongside their resulting visual errors, the agent can systematically converge toward the exact target state. For instance, to rectify a spatial collision (\eg, a basketball intersecting a table), the agent actively contrasts prior height parameters (0.8 vs. 0.7) to deduce the precise geometric configuration (0.75). Without this evolving memory, the agent is restricted to isolated, single-step edits, inevitably oscillating between incorrect states rather than converging on exact parameters.

\noindent \textbf{Mitigating Context Bloat.} 
While memory ensures continuous refinement, naively accumulating the complete multimodal trajectory introduces a critical bottleneck. Such unbounded expansion causes context bloat, scaling the computational overhead and introducing distracting noise---a phenomenon known as \textit{context rot}~\citep{hong2025context, liu-etal-2024-lost}. As the sequence lengthens, the foundation model's reasoning precision drastically degrades. To combat this, we constrain the memory using a fixed-size sliding window that retains only the most recent $L$ iterations. \cref{fig:memory_ablation} demonstrates this necessity: while our window-controlled memory maintains stable generation quality over long horizons, an unconstrained memory baseline rapidly collapses. Crucially, this aggressive truncation is inherently lossless. Because the latest synthesized program $p_t$ functions as a self-contained symbolic state that strictly encapsulates all prior successful geometric edits, it inherently renders older, intermediate reasoning traces redundant.

\begin{table}[t]
    \centering
    \scriptsize
    \setlength{\tabcolsep}{3pt}
    \renewcommand{\arraystretch}{0.95}
    \begin{tabularx}{\linewidth}{@{} l l >{\raggedright\arraybackslash}X @{}}
    \toprule
    \textbf{Semantic Skill} & \textbf{Encapsulated APIs} & \textbf{Capability} \\
    \midrule
    \multicolumn{3}{@{}l}{\textbf{\textit{State Observation}}} \\
    Spatial Navigation & \texttt{set\_camera}, \texttt{investigate} & Actively adjust viewpoints to inspect regions and bypass occlusions. \\
    Temporal Inspection & \texttt{set\_keyframe} & Navigate the 4D timeline to evaluate dynamic physical motions. \\
    Structural Query & \texttt{get\_scene\_info}, \texttt{set\_visibility} & Query geometric attributes and toggle object visibility. \\
    Diagnostic Rendering & \texttt{initialize\_view} & Render canonical views and compute target bounding boxes. \\
    \midrule
    \multicolumn{3}{@{}l}{\textbf{\textit{State Modification}}} \\
    Asset Instantiation & \texttt{init\_scene}, \texttt{get\_better\_assets} & Retrieve or generate high-fidelity 3D assets via external modules to populate scenes. \\
    State Execution & \texttt{execute\_code} & Apply synthesized programs to persistently update scene states. \\
    Trajectory Control & \texttt{undo\_action}, \texttt{end\_process} & Revert erroneous edits or cleanly terminate the reasoning loop. \\
    \bottomrule
    \end{tabularx}
    \vspace{4pt}
    \caption{\textbf{High-Level Skill Library as Semantic Interfaces.} Rather than exposing raw graphics APIs, we equip the agent with encapsulated capabilities. The library is bifurcated into State Observation (gathering visual/structural context without altering the scene) and State Modification (executing persistent changes), mirroring our dual-space reasoning framework.}
    \label{tab:skill-library}
    \vspace{-4mm}
\end{table}

\subsection{Skill Library as a Semantic Interface for Graphics}
\label{sec:3.3:skills}

\noindent \textbf{Action Space Compression.} 
A naive approach to scene generation is to prompt the VLM to output raw, low-level graphics API calls (\eg, raw Blender Python scripts). However, standard graphics engines feature a high-dimensional and brittle action space. Exposing the agent directly to this low-level syntax severely exacerbates context bloat and frequently leads to syntactic hallucinations or non-executable loops. To address this, we conceptualize our skill library not as rigid low-level APIs, but as a set of high-level semantic interfaces (\cref{tab:skill-library}). By encapsulating fragile, atomic operations into reliable, macro-level skills, we significantly compress the action space. This structural abstraction essentially insulates the agent from low-level programming trivialities, freeing its entire context window and cognitive capacity for what it excels at: high-level multimodal reasoning and abstract planning~\citep{anthropic2026skills}.

\noindent \textbf{Perception, Reasoning, and Action.} 
To seamlessly bridge the symbolic and visual modalities, the agent operates through a continuous triad of perception, reasoning, and action. At the core of this loop is the agent's internal reasoning capacity, which cautiously orchestrates the external skill library (\cref{tab:skill-library}). Rather than blindly manipulating the environment, the agent first relies on \textit{State Observation} skills (\eg, \texttt{investigate} and \texttt{get\_scene\_info}). Because these interfaces are inherently read-only and non-destructive, the agent is free to conduct exhaustive, multi-turn spatial queries to achieve fine-grained visual grounding. Crucially, this unconstrained exploration facilitates a cognitive leap from visual appearance to structural abstraction. It allows the VLM to distill raw visual phenomena into a high-level semantic understanding of the scene's underlying physical, geometric, and spatial properties. Only after this rigorous transition from surface-level grounding to abstract reasoning does the agent trigger a persistent change via \textit{State Modification} skills. Central to this is \texttt{execute\_code}, which translates this semantic understanding into exact parametric adjustments by injecting the synthesized program ($p_t$) into the engine. By strictly decoupling fine-grained visual perception from abstract code execution, the framework ensures that scene mutations are conditioned on a comprehensive series of prior observations and are highly deliberate.

\section{Experiments}
\label{sec:4:experiments}

We evaluate \ours{} across a diverse set of visual generation and editing tasks to demonstrate the effectiveness of its interleaved multimodal reasoning paradigm. 
We start by presenting qualitative results that highlight the visual fidelity and broad adaptability of \ours{} across 2D, 3D, and 4D domains (\cref{sec:4.1:qualitative_results}).
Recognizing that traditional benchmarks primarily assess static, single-turn outputs, we introduce BlenderBench, a challenging new benchmark designed to stress-test multi-step, dynamic agentic behaviors (\cref{sec:4.2:blenderbench}). 
Furthermore, we demonstrate the generalization capability of \ours{}, achieving substantial performance gains on established quantitative benchmarks (\cref{sec:4.3:standard_benchmarks}).

\begin{figure}[t]
    \centering
    \begin{minipage}[t]{0.59\linewidth}
        \vspace{0pt}
        \centering
        \includegraphics[width=\linewidth]{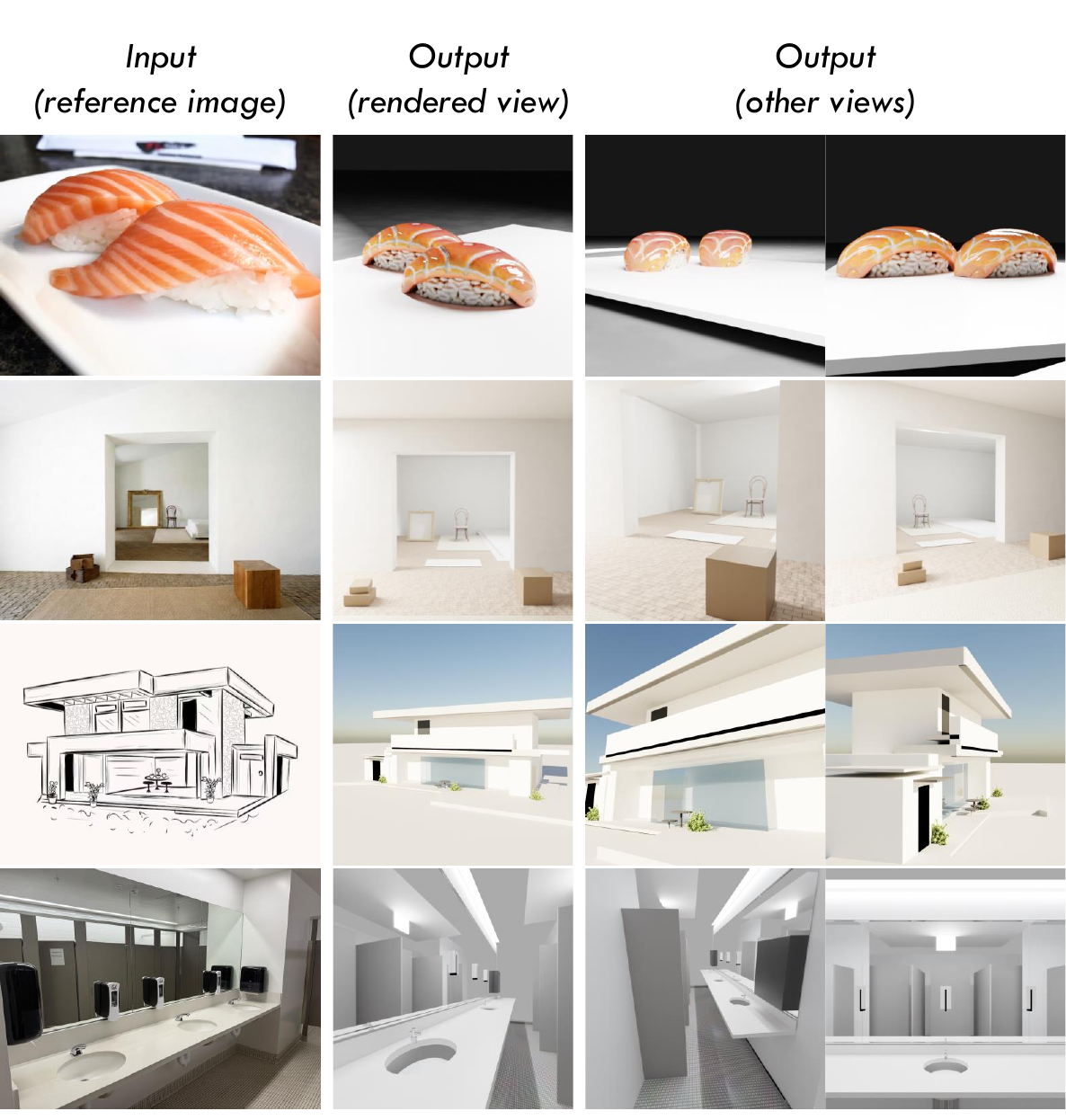}
    \end{minipage}\hfill
    \begin{minipage}[t]{0.39\linewidth}
        \vspace{0pt}
        \centering
        \includegraphics[width=\linewidth]{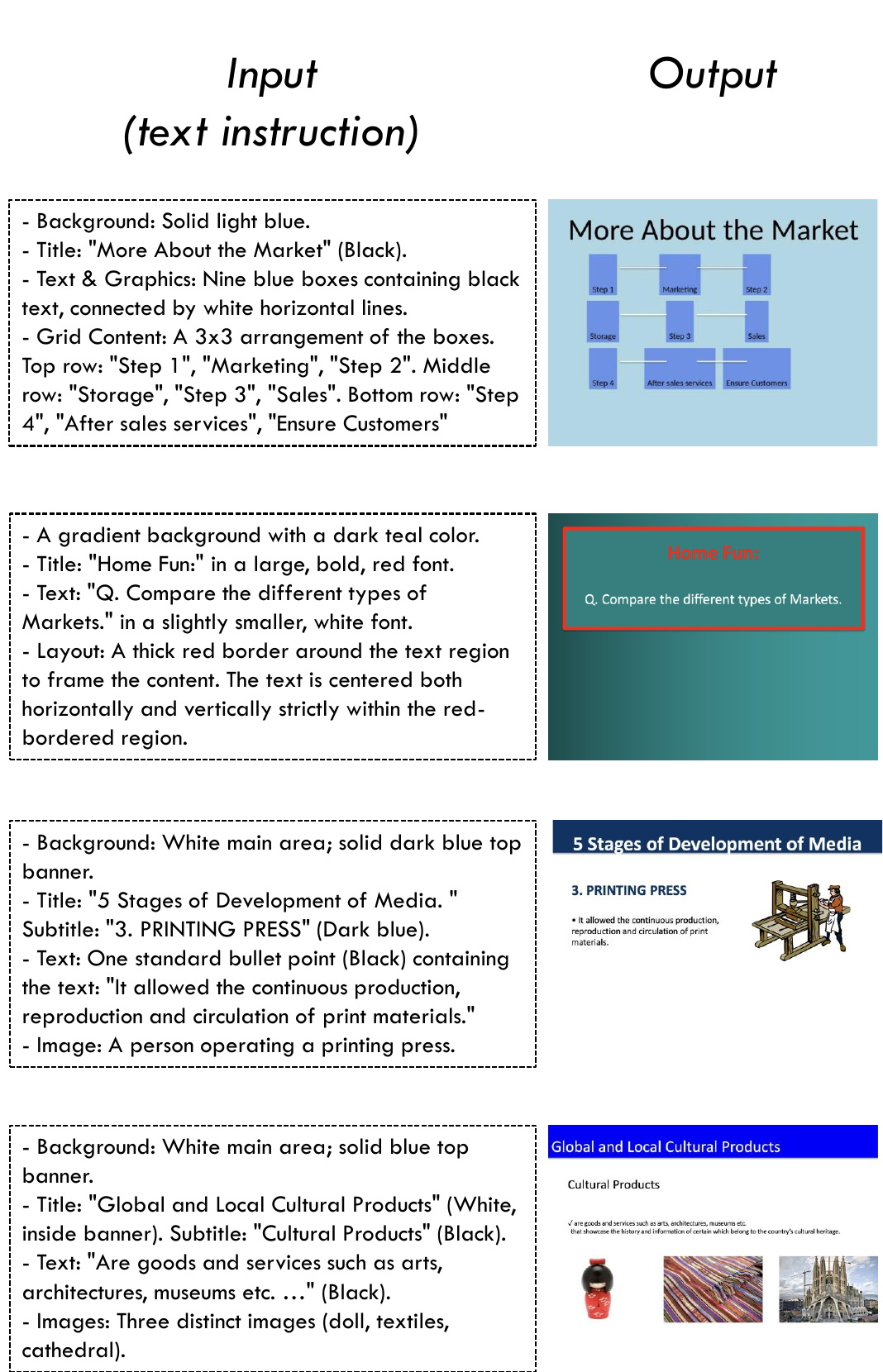}
    \end{minipage}
    
    \vspace{0.5em} 
    \begin{minipage}[t]{0.59\linewidth}
        \caption{\textbf{Qualitative Results on 3D Scene Generation.} \ours accurately generates high-fidelity 3D scenes from diverse visual inputs with precise semantic and visual alignment.}
        \label{fig:demo_3d}
    \end{minipage}\hfill
    \begin{minipage}[t]{0.39\linewidth}
        \caption{\textbf{Qualitative Results on 2D Document Design.} \ours generates high-quality presentation slides from text instructions.}
        \label{fig:demo_2d}
    \end{minipage}
\end{figure}

\begin{figure}[t]
    \centering
    \includegraphics[width=1\linewidth]{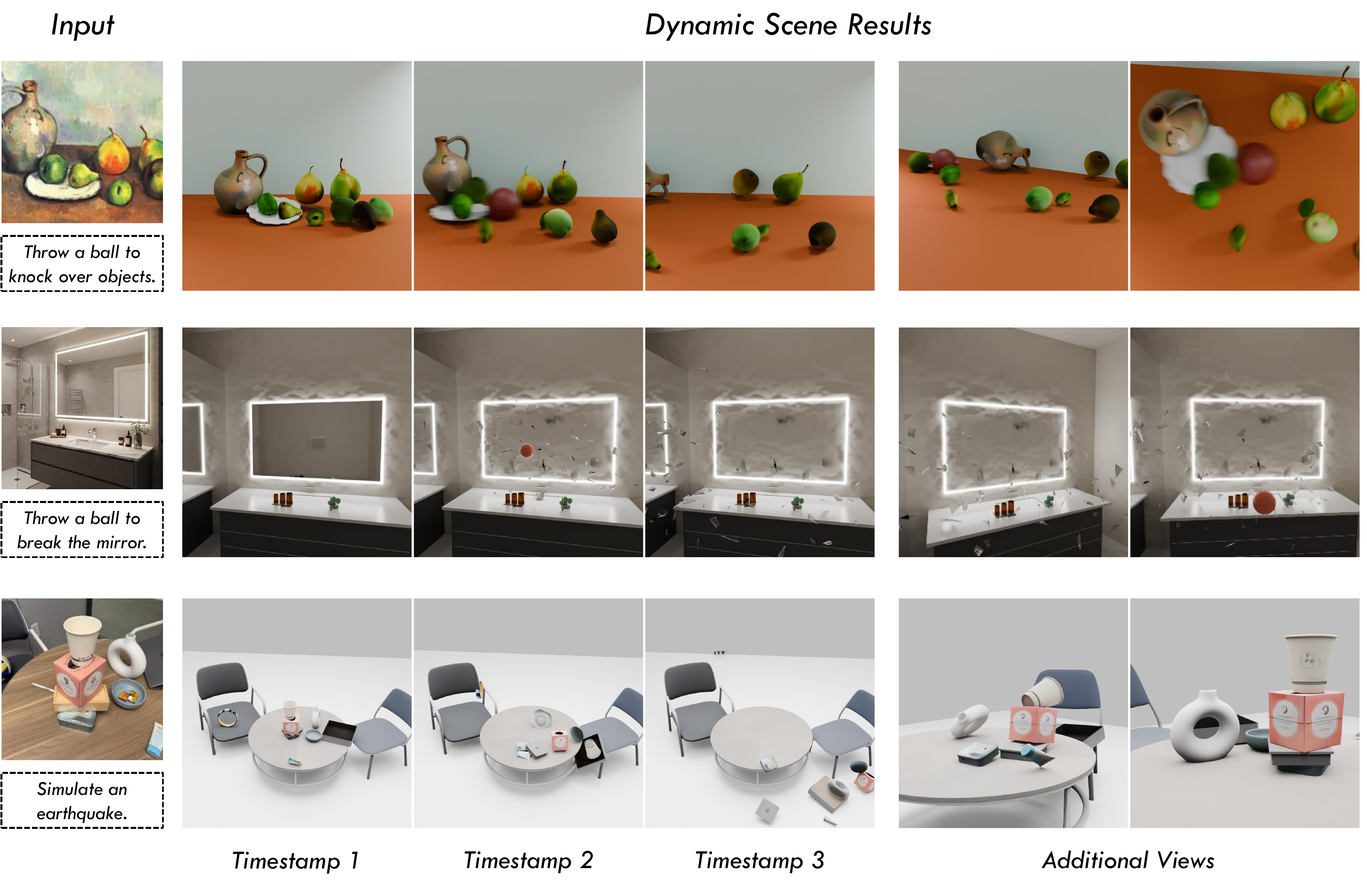}
    \caption{\textbf{Qualitative Results on 4D Dynamic Scene Simulation.} Given a single reference image and a text instruction, \ours{} accurately synthesizes complex physical dynamics (\eg, multi-object collisions and glass shattering). It adaptively constructs scenes by fetching SAM-3D~\citep{sam3dteam2025sam3d3dfyimages} assets via \texttt{init\_scene} (\eg, fruit, desk) or generating them programmatically from primitives (\eg, mirror).}
    \label{fig:demo_4d}
\end{figure}

\subsection{Qualitative Results across Diverse Tasks}
\label{sec:4.1:qualitative_results}

We present qualitative results across 3D scene reconstruction, 4D physics simulation, and 2D document design to demonstrate the generation capabilities and broad adaptability of \ours{}. These examples illustrate that a unified interleaved reasoning loop naturally generalizes to diverse visual tasks without relying on domain-specific architectures or heuristics.

\noindent \textbf{3D Generation and Editing.} 
As illustrated in \cref{fig:demo_3d}, \ours{} reconstructs diverse 3D scenes from sparse single-view inputs, ranging from photorealistic images to hand-drawn sketches. When populating scenes spanning from object-centric arrangements to complex indoor environments (\eg, sushi, bedroom), the agent dynamically invokes the \texttt{get\_better\_assets} tool to retrieve and position high-quality external meshes. It then iteratively optimizes the spatial layout and material properties of these assets to faithfully reproduce the target's lighting and textures. Beyond parsing natural images, \ours{} exhibits strong sketch-to-3D generalization, converting sparse line drawings (\eg, an architectural house sketch) into complete 3D geometries.

\noindent \textbf{4D Dynamic Scene Simulation.} 
Beyond static reconstruction, \ours{} simulates 4D physical dynamics given a single image and a text prompt (\cref{fig:demo_4d}). By inferring and configuring underlying physical parameters (\eg, mass and collision shapes), the agent translates static scenes into complex interactive environments, such as a ball scattering fruit, a mirror shattering into reflective shards, or an earthquake disrupting tabletop objects. While maintaining consistent reflections on moving glass or resolving multi-object collisions remains a fundamental bottleneck for pixel-based generative models, \ours{} bypasses these limitations. By executing programmatic commands within a physics engine, the framework inherently enforces strict physical accuracy and temporal consistency.

\noindent \textbf{2D Document Design.} 
In addition to 3D and 4D spatial graphics, \cref{fig:demo_2d} demonstrates the framework's applicability to 2D document generation. By substituting the underlying execution engine, \ours{} synthesizes professional PowerPoint slides from scratch, orchestrating layout planning, text formatting, and image placement. This flexibility underscores a core advantage of our approach: interleaved multimodal reasoning naturally generalizes to any visual domain governed by a programmatic interface, utilizing this structured action space to iteratively converge on the target state.

\begin{figure}[t]
    \centering
    \includegraphics[width=\linewidth]{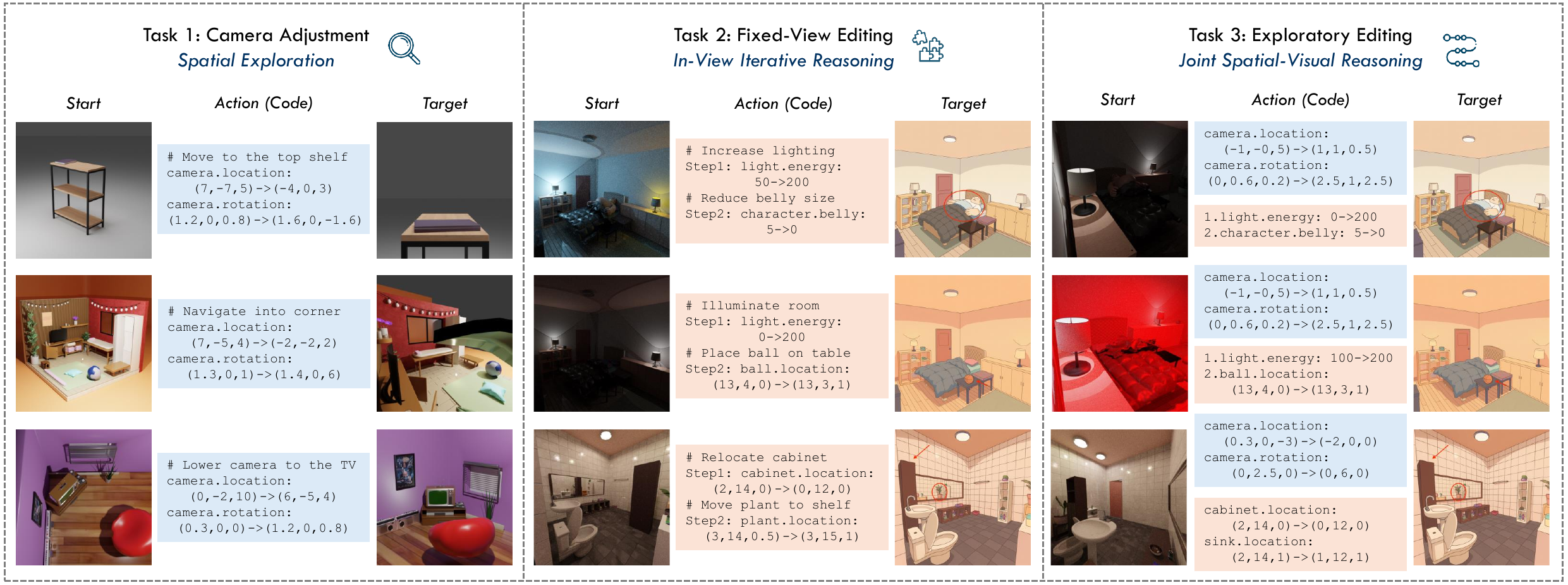} 
    \caption{\textbf{Overview of the BlenderBench evaluation tracks.} The benchmark comprises three progressive scenarios to stress-test interleaved multimodal reasoning. \textit{Task 1 (Camera Adjustment)} isolates active spatial exploration for viewpoint alignment. \textit{Task 2 (Fixed-View Editing)} evaluates iterative scene reasoning and attribute modification under a fixed camera pose. \textit{Task 3 (Exploratory Editing)} unifies these challenges, requiring the agent to actively explore the spatial layout while simultaneously executing programmatic multi-step edits.}
    \label{fig:blenderbench}
\end{figure}

\begin{table}[t]
\centering

\small
\resizebox{\textwidth}{!}{
\begin{tabular}{l l cc cc cc cc cc c}
\toprule
\multirow{2}{*}{Model} & \multirow{2}{*}{Setting} 
& \multicolumn{3}{c}{Task 1} 
& \multicolumn{3}{c}{Task 2} 
& \multicolumn{3}{c}{Task 3} & \multirow{2}{*}{\makecell{Impr.\\(\%)}}\\
\cmidrule(lr){3-5} \cmidrule(lr){6-8} \cmidrule(lr){9-11}
& & PL $\downarrow$ & N-CLIP $\downarrow$ & VLM $\uparrow$
  & PL $\downarrow$ & N-CLIP $\downarrow$ & VLM $\uparrow$
  & PL $\downarrow$ & N-CLIP $\downarrow$ & VLM $\uparrow$ \\
\midrule
\multirow{7}{*}{GPT-4o} 
& One-Shot & 48.16 & 64.17 & 0.58 & 7.36 & 7.12 & 2.75 & 30.14 & 38.69 & 0.25 & - \\
\cmidrule(lr){2-12}
& BlenderAlchemy (best-of-1) & 10.62 & \textbf{17.68} & \textbf{1.75} 
& 6.13 & 7.71 & 3.08 
& 19.60 & 26.10 & 0.67 
& 67.56 \\
& \ours (best-of-1) & \textbf{8.56} & 18.19 & 1.44 
& \textbf{5.11} & \textbf{4.33} & \textbf{3.58}
& \textbf{14.51} & \textbf{17.91} & \textbf{1.53}
& \textbf{113.96} \\
& 
& \textcolor{darkgreen}{(+19.40\%)} 
& \textcolor{red}{(-2.89\%)} 
& \textcolor{red}{(-17.71\%)} 
& \textcolor{darkgreen}{(+16.64\%)}  
& \textcolor{darkgreen}{(+43.84\%)} 
& \textcolor{darkgreen}{(+16.23\%)} 
& \textcolor{darkgreen}{(+25.97\%)}  
& \textcolor{darkgreen}{(+31.40\%)} 
& \textcolor{darkgreen}{(+128.36\%)} 
& \textcolor{darkgreen}{(+68.69\%)} \\
\cmidrule(lr){2-12}
& BlenderAlchemy (best-of-4) & 14.50 & 19.57 & 1.75
& \textbf{1.95} & \textbf{2.47} & 3.53
& 20.62 & 25.79 & 0.56
& 77.48 \\
& \ours (best-of-4) & \textbf{5.47} & \textbf{6.10} & \textbf{3.25}
& 2.94 & 3.50 & \textbf{3.83}
& \textbf{12.62} & \textbf{22.84} & \textbf{1.61}
& \textbf{159.19} \\
&
& \textcolor{darkgreen}{(+62.28\%)} 
& \textcolor{darkgreen}{(+68.84\%)} 
& \textcolor{darkgreen}{(+85.71\%)} 

& \textcolor{red}{(-50.77\%)} 
& \textcolor{red}{(-41.70\%)} 
& \textcolor{darkgreen}{(+8.50\%)} 

& \textcolor{darkgreen}{(+38.79\%)} 
& \textcolor{darkgreen}{(+11.43\%)} 
& \textcolor{darkgreen}{(+187.50\%)} 

& \textcolor{darkgreen}{(+105.45\%)} \\
\midrule
\multirow{7}{*}{Qwen3-VL-8B} 
& One-Shot & 60.82 & 78.16 & 0.28 & 33.14 & 37.51 & 1.61 & 22.81 & 22.98 & 1.25 & - \\
\cmidrule(lr){2-12}
& BlenderAlchemy (best-of-1) & 25.61 & 37.93 & 0.36 & 10.69 & 13.92 & 2.64 & 30.45 & 23.91 & 0.64 & 27.36 \\
& \ours (best-of-1) & \textbf{10.54} & \textbf{17.27} & \textbf{1.31} & \textbf{3.85} & \textbf{3.34} & \textbf{3.33} & \textbf{11.35} & \textbf{6.86} & \textbf{2.25} & \textbf{112.79} \\
&
& \textcolor{darkgreen}{(+58.86\%)} 
& \textcolor{darkgreen}{(+54.45\%)} 
& \textcolor{darkgreen}{(+263.89\%)} 

& \textcolor{darkgreen}{(+63.98\%)} 
& \textcolor{darkgreen}{(+76.06\%)} 
& \textcolor{darkgreen}{(+26.14\%)} 

& \textcolor{darkgreen}{(+62.75\%)} 
& \textcolor{darkgreen}{(+71.33\%)} 
& \textcolor{darkgreen}{(+251.56\%)} 

& \textcolor{darkgreen}{(+312.20\%)} \\
\cmidrule(lr){2-12}
& BlenderAlchemy (best-of-4) & 11.57 & 17.58 & 1.19
& 5.38 & 4.64 & 2.94
& 23.62 & 15.67 & 0.93
& 82.24 \\
& \ours (best-of-4) & \textbf{8.80} & \textbf{15.67} & \textbf{1.38}
& \textbf{5.02} & \textbf{3.86} & \textbf{3.08}
& \textbf{9.08} & \textbf{6.94} & \textbf{2.02}
& \textbf{112.87}\\
&
& \textcolor{darkgreen}{(+23.92\%)} 
& \textcolor{darkgreen}{(+10.88\%)} 
& \textcolor{darkgreen}{(+15.97\%)} 

& \textcolor{darkgreen}{(+6.69\%)} 
& \textcolor{darkgreen}{(+16.81\%)} 
& \textcolor{darkgreen}{(+4.76\%)} 

& \textcolor{darkgreen}{(+61.54\%)} 
& \textcolor{darkgreen}{(+55.73\%)} 
& \textcolor{darkgreen}{(+117.20\%)} 

& \textcolor{darkgreen}{(+37.20\%)} \\
\bottomrule
\end{tabular}
}
\captionsetup{aboveskip=6pt, belowskip=0pt}
\caption{\textbf{Evaluation on BlenderBench.} We report Photometric Loss (PL), Negative-CLIP Score (N-CLIP), and VLM Score (VLM).}
\label{tab:quantitative_blenderbench}
\end{table}

\subsection{BlenderBench: Evaluating Agentic Behaviors}
\label{sec:4.2:blenderbench}

Inspired by prior works on single-step graphics editing (\eg, BlenderGym~\citep{gu2025blendergym}), we introduce BlenderBench, a new benchmark designed to stress-test interleaved multimodal reasoning. As illustrated in \cref{fig:blenderbench}, while traditional benchmarks primarily evaluate static final outputs, BlenderBench explicitly quantifies multi-step, dynamic agentic behaviors, such as active spatial exploration, iterative scene reasoning for attribute modification, and complex exploratory editing.

\noindent \textbf{Setup.}
To stress-test interleaved multimodal reasoning, BlenderBench comprises three progressive tracks: Task 1 (Camera Adjustment), Task 2 (Fixed-View Editing), and Task 3 (Exploratory Editing). Because target references frequently incorporate domain gaps (\eg, style transfers), we evaluate generation quality using Photometric Loss (PL) and Negative-CLIP Score (N-CLIP), and introduce a \textit{VLM Score} to prioritize structural layout over task-irrelevant stylistic artifacts. Finally, we compare \ours{} against a standard one-shot approach and the memory-less baseline BlenderAlchemy~\citep{huang2024blenderalchemy}. To ensure fair iterative comparisons, we adopt a \textit{best-of-$N$} setting~\citep{gu2025blendergym} to analyze the impact of search budgets. In this setting, the agent generates $N$ candidate edits at each step and selects the optimal one for the next round. 

\noindent \textbf{Results.} 
As shown in \cref{tab:quantitative_blenderbench}, \ours{} consistently surpasses existing baselines, yielding an average relative improvement (Impr.) of +124.70\% across all evaluation tracks and model settings. In Task 1 (Camera Adjustment), standard one-shot methods fail due to limited zero-shot spatial perception, whereas our agent dynamically updates its viewpoint to locate targets. Notably, scaling the \textit{best-of-$N$} search budget from $N=1$ to $N=4$ under GPT-4o boosts the VLM Score from 1.44 to 3.25, effectively overcoming local minima during viewpoint alignment.
Furthermore, for joint spatial-visual reasoning in Task 3 (Exploratory Editing), BlenderAlchemy suffers from cross-step inconsistency, frequently overwriting earlier edits. By explicitly tracking states, \ours{} executes complex modifications while strictly maintaining sequential consistency, achieving a +187.50\% VLM Score improvement (GPT-4o, best-of-4). 
Beyond absolute metrics, the iterative reasoning loop acts as a critical scaffold across varying model capacities. When instantiated with Qwen, \ours{} delivers a +312.20\% relative gain over BlenderAlchemy (best-of-1), effectively bridging the performance gap between lightweight open-source models and larger closed-source models.

\begin{table}[t]
    \centering
    \small
    
    \begin{minipage}[t]{0.69\linewidth}
        \vspace{0pt}
        \centering
        \captionsetup{aboveskip=6pt, belowskip=0pt}
        \setlength{\tabcolsep}{2.5pt}
        \resizebox{\linewidth}{!}{%
            \begin{tabular}{@{} l l cc cc cc cc cc c @{}}
                \toprule
                \multirow{2}{*}{Model} & \multirow{2}{*}{Setting} 
                & \multicolumn{2}{c}{Shape} 
                & \multicolumn{2}{c}{Place} 
                & \multicolumn{2}{c}{Geom} 
                & \multicolumn{2}{c}{Light} 
                & \multicolumn{2}{c}{Mat} & \multirow{2}{*}{\makecell{Impr.\\(\%)}}\\
                \cmidrule(lr){3-4} \cmidrule(lr){5-6} \cmidrule(lr){7-8} \cmidrule(lr){9-10} \cmidrule(lr){11-12}
                & & PL $\downarrow$ & NC $\downarrow$ 
                  & PL $\downarrow$ & NC $\downarrow$ 
                  & PL $\downarrow$ & NC $\downarrow$ 
                  & PL $\downarrow$ & NC $\downarrow$ 
                  & PL $\downarrow$ & NC $\downarrow$ \\
                \midrule
                \multirow{7}{*}{GPT-4o} 
                & One-Shot & 7.94 & 19.96 & 11.86 & 29.31 & 18.12 & 24.91 & 2.06 & 2.17 & 8.78 & 15.25 & -  \\
                \cmidrule(lr){2-13}
                & B-Alchemy (bo1) & 7.82 & 20.19 & 11.57 & 28.07 & 15.58 & 20.61 & 1.44 & 1.82 & 7.22 & 12.05 & 12.33 \\
                & \ours (bo1)
                & \textbf{6.70} & \textbf{16.70} & \textbf{9.89} & \textbf{26.87} & \textbf{11.82} & \textbf{15.76} & \textbf{1.43} & \textbf{1.41} & \textbf{6.11} & \textbf{9.39} & \textbf{26.88} \\
                & & \scriptsize\textcolor{darkgreen}{(+14.3\%)} 
                & \scriptsize\textcolor{darkgreen}{(+17.3\%)} 
                & \scriptsize\textcolor{darkgreen}{(+14.5\%)} 
                & \scriptsize\textcolor{darkgreen}{(+4.3\%)}  
                & \scriptsize\textcolor{darkgreen}{(+24.1\%)} 
                & \scriptsize\textcolor{darkgreen}{(+23.5\%)} 
                & \scriptsize\textcolor{darkgreen}{(+0.7\%)}  
                & \scriptsize\textcolor{darkgreen}{(+22.5\%)} 
                & \scriptsize\textcolor{darkgreen}{(+15.4\%)} 
                & \scriptsize\textcolor{darkgreen}{(+22.1\%)} 
                & \scriptsize\textcolor{darkgreen}{(+118.1\%)} \\
                \cmidrule(lr){2-13}
                & B-Alchemy (bo4) & 5.78 & 17.19 & 10.29 & 25.58 & 9.47 & 11.68 & 0.97 & 1.23 & 9.47 & 12.20 & 27.63 \\
                & \ours (bo4) & \textbf{5.45} & \textbf{13.08} & \textbf{9.76} & \textbf{17.48} & \textbf{7.38} & \textbf{7.36} & \textbf{0.70} & \textbf{1.01} & \textbf{3.04} & \textbf{7.63} & \textbf{48.86} \\
                & 
                & \scriptsize\textcolor{darkgreen}{(+5.7\%)} 
                & \scriptsize\textcolor{darkgreen}{(+23.9\%)} 
                & \scriptsize\textcolor{darkgreen}{(+5.1\%)} 
                & \scriptsize\textcolor{darkgreen}{(+31.7\%)}  
                & \scriptsize\textcolor{darkgreen}{(+22.1\%)} 
                & \scriptsize\textcolor{darkgreen}{(+37.0\%)} 
                & \scriptsize\textcolor{darkgreen}{(+27.8\%)}  
                & \scriptsize\textcolor{darkgreen}{(+17.9\%)} 
                & \scriptsize\textcolor{darkgreen}{(+67.9\%)} 
                & \scriptsize\textcolor{darkgreen}{(+37.5\%)} 
                & \scriptsize\textcolor{darkgreen}{(+76.8\%)} \\
                \midrule
                \multirow{7}{*}{Qwen3-8B} 
                & One-Shot & 7.76 & 21.78 & 14.10 & 41.44 & 11.65 & 14.02 & 12.12 & 12.30 & 9.83 & 17.16 & - \\
                \cmidrule(lr){2-13}
                & B-Alchemy (bo1) & \textbf{12.11} & 31.50 & 13.86 & 39.47 & 9.62 & 14.47 & 5.22 & 6.44 & 6.62 & 16.98 & 5.81 \\
                & \ours (bo1) & 13.51 & \textbf{27.07} & \textbf{10.07} & \textbf{28.12} & \textbf{9.15} & \textbf{12.24} & \textbf{2.33} & \textbf{2.58} & \textbf{6.43} & \textbf{11.53} & \textbf{22.64} \\
                & 
                & \scriptsize\textcolor{red}{(-11.6\%)} 
                & \scriptsize\textcolor{darkgreen}{(+14.1\%)} 
                & \scriptsize\textcolor{darkgreen}{(+27.3\%)} 
                & \scriptsize\textcolor{darkgreen}{(+28.8\%)}  
                & \scriptsize\textcolor{darkgreen}{(+4.9\%)} 
                & \scriptsize\textcolor{darkgreen}{(+15.4\%)} 
                & \scriptsize\textcolor{darkgreen}{(+55.4\%)}  
                & \scriptsize\textcolor{darkgreen}{(+59.9\%)} 
                & \scriptsize\textcolor{darkgreen}{(+2.9\%)} 
                & \scriptsize\textcolor{darkgreen}{(+32.1\%)} 
                & \scriptsize\textcolor{darkgreen}{(+289.8\%)} \\
                \cmidrule(lr){2-13}
                & B-Alchemy (bo4) & 9.20 & 19.60 & 13.07 & 36.34 & 9.24 & 12.62 & 3.68 & 4.23 & 4.42 & 17.12 & 23.23 \\
                & \ours (bo4) & \textbf{7.20} & \textbf{17.97} & \textbf{9.92} & \textbf{24.28} & \textbf{8.69} & \textbf{10.08} & \textbf{1.89} & \textbf{2.53} & \textbf{3.06} & \textbf{9.30} & \textbf{42.88} \\
                & 
                & \scriptsize\textcolor{darkgreen}{(+21.7\%)} 
                & \scriptsize\textcolor{darkgreen}{(+8.3\%)} 
                & \scriptsize\textcolor{darkgreen}{(+24.1\%)} 
                & \scriptsize\textcolor{darkgreen}{(+33.2\%)}  
                & \scriptsize\textcolor{darkgreen}{(+6.0\%)} 
                & \scriptsize\textcolor{darkgreen}{(+20.1\%)} 
                & \scriptsize\textcolor{darkgreen}{(+48.6\%)}  
                & \scriptsize\textcolor{darkgreen}{(+40.2\%)} 
                & \scriptsize\textcolor{darkgreen}{(+30.8\%)} 
                & \scriptsize\textcolor{darkgreen}{(+45.7\%)} 
                & \scriptsize\textcolor{darkgreen}{(+84.6\%)} \\
                \bottomrule
            \end{tabular}%
        }
        \caption{\textbf{Evaluation on BlenderGym.} We report Photometric Loss (PL) and Negative-CLIP Score (NC). Here bo$N$ denotes the best-of-$N$ setting.}
        \label{tab:quantitative_blendergym}
    \end{minipage}\hfill
    \begin{minipage}[t]{0.29\linewidth}
        \vspace{0pt}
        \centering
        \captionsetup{aboveskip=6pt, belowskip=0pt}
        \setlength{\tabcolsep}{3pt}
        
        \renewcommand{\arraystretch}{1} 
        
        \resizebox{\linewidth}{!}{%
            \begin{tabular}{@{} l l ccc @{}}
                \toprule
                {Model} & {Setting} & {Exec. $\uparrow$} & {Qual. $\uparrow$} & {Over. $\uparrow$} \\
                \midrule
                \multirow{3}{*}{GPT-4o}
                & One-Shot & 0.92 & 60.7 & 55.8  \\
                \cmidrule(lr){2-5}
                & \ours & \textbf{0.95} & \textbf{61.4} & \textbf{58.3} \\
                & & \textcolor{darkgreen}{(+3.3\%)} & \textcolor{darkgreen}{(+1.2\%)} & \textcolor{darkgreen}{(+4.5\%)} \\
                \midrule
                \multirow{3}{*}{Claude-3.5}
                & One-Shot & \textbf{0.93} & 65.3 & 60.7 \\
                \cmidrule(lr){2-5}
                & \ours & 0.90 & \textbf{69.2} & \textbf{62.2} \\
                & & \textcolor{red}{(-3.2\%)} & \textcolor{darkgreen}{(+6.0\%)} & \textcolor{darkgreen}{(+2.5\%)} \\
                \midrule
                \multirow{3}{*}{Gemini-2.5}
                & One-Shot & 0.61 & 67.9 & 41.4  \\
                \cmidrule(lr){2-5}
                & \ours & \textbf{0.69} & \textbf{70.9} & \textbf{48.9} \\
                & & \textcolor{darkgreen}{(+13.1\%)} & \textcolor{darkgreen}{(+4.4\%)} & \textcolor{darkgreen}{(+18.1\%)} \\
                \midrule
                \multirow{3}{*}{Qwen3-8B}
                & One-Shot & 0.10 & 60.5 & 6.0 \\
                \cmidrule(lr){2-5}
                & \ours & \textbf{0.54} & \textbf{60.8} & \textbf{32.8} \\
                & & \textcolor{darkgreen}{(+440\%)} & \textcolor{darkgreen}{(+0.5\%)} & \textcolor{darkgreen}{(+446.7\%)} \\
                \bottomrule
            \end{tabular}%
        }
        \caption{\textbf{Evaluation on SlideBench.} We report Execution, Quality, and Overall scores.}
        \label{tab:quantitative_slidebench}
    \end{minipage}
\end{table}

\subsection{Evaluation on Existing Benchmarks}
\label{sec:4.3:standard_benchmarks}

While primarily developed for interleaved multimodal reasoning, \ours{} naturally generalizes to standard single-step programmatic generation and editing tasks. To demonstrate this strong task-agnostic capability, we evaluate the framework across two established benchmarks from distinct visual domains.

\noindent \textbf{3D Editing (BlenderGym).} 
We first evaluate \ours{} on BlenderGym~\citep{gu2025blendergym} for single-step 3D scene editing. As reported in~\cref{tab:quantitative_blendergym}, \ours{} consistently surpasses both the one-shot baseline and the memory-less baseline BlenderAlchemy~\citep{huang2024blenderalchemy}. \ours{} yields an average improvement (Impr.) of +35.32\% across all the settings. These results indicate that \ours precisely localizes and modifies fine-grained object attributes (\eg, blend shapes, placement, and lighting) while strictly preserving the unedited elements of the scene.

\noindent \textbf{2D Design (SlideBench).} 
Extending to 2D environments, we evaluate \ours{} on SlideBench~\citep{ge2025autopresent} for programmatic PowerPoint generation. As reported in~\cref{tab:quantitative_slidebench}, \ours{} delivers an average Overall score improvement of +117.17\%  across a diverse spectrum of foundation models, spanning open-source (\eg, Qwen3-VL-8B) and closed-source (\eg, GPT-4o, Claude-Sonnet-4) models. While standard one-shot methods frequently halt upon encountering a single syntax or rendering error, our interleaved reasoning loop acts as a robust error-recovery mechanism. By iteratively parsing execution logs to rectify faulty code, \ours{} drastically improves the execution success rate (Exec.), ensuring the reliable generation of complex document layouts.

\section{Conclusion}

We present \ours{}, a multimodal agent that tackles inverse graphics through interleaved multimodal reasoning. By deeply coupling discrete program synthesis with active visual verification, \ours{} maps abstract spatial concepts to precise, executable code across 2D, 3D, and 4D domains. To systematically assess these dynamic agentic behaviors, we introduce BlenderBench, a comprehensive benchmark designed to quantify multi-step spatial reasoning and progressive scene editing. While current performance is bounded by the spatial perception capabilities of underlying VLMs and context window constraints in extremely long sequences, our framework is inherently extensible. As foundation models and external generative tools (\eg, Meshy~\citep{meshy_ai}, Tripo~\citep{tripo_ai}, SAM-3D~\citep{sam3dteam2025sam3d3dfyimages}) advance, their capabilities will seamlessly plug into our closed-loop paradigm. Ultimately, by orchestrating an evolving memory and a semantic skill library into a unified, training-free agent, \ours{} tackles complex, long-horizon inverse graphics, serving as a foundational step toward fully automated scene generation.

\bibliographystyle{splncs04}
\bibliography{main}

\appendix
\clearpage
\thispagestyle{empty}

\begin{center}
    {\Large \bfseries Vision-as-Inverse-Graphics Agent \\[2mm] via Interleaved Multimodal Reasoning \par} 
    \vspace{4mm}
    {\large \bfseries Supplementary Material \par}
\end{center}

\section{Evaluation Settings}
\label{sec:appendix_eval_settings}

\subsection{Quantitative Settings}

\paragraph{Implementation Details.}
For the baseline one-shot setting, we prompt the VLMs using the standard system prompts from~\citep{ge2025autopresent, gu2025blendergym} to ensure a fair comparison. For \ours{}, we set \texttt{max\_iterate\_round=10} to prevent excessive iterations beyond the search budget. During the \textit{best-of-$N$} evaluation, the agent explores multiple reasoning branches to generate a set of candidate modifications. We employ a rule-based selection mechanism that evaluates these candidates via their CLIP score, propagating the best-aligned geometric state to the next iteration. For tool invocation of the VLMs, we leverage the native \texttt{tool\_calls} API for the GPT, Claude, and Gemini model families. Conversely, as the open-source Qwen model lacks a highly optimized native tool-calling interface, we prompt it to directly output arguments in a strict JSON format. This explicitly bypasses brittle API wrappers and effectively mitigates formatting errors.

\paragraph{BlenderBench.}
To stress-test interleaved multimodal reasoning, BlenderBench comprises three progressive evaluation tracks: Task 1 (Camera Adjustment), Task 2 (Fixed-View Editing), and Task 3 (Exploratory Editing). Crucially, the target reference images frequently incorporate domain gaps, such as artistic style transfers and explicit visual prompts (\eg, guiding arrows). This forces the agent to extract geometric intent rather than overfit to pixel-level patterns. We assess standard generation quality using Photometric Loss (PL) and Negative-CLIP Score (N-CLIP). Furthermore, to explicitly quantify task success under these stylistic shifts, we introduce the \textit{VLM Score}. It assesses outputs across four fine-grained dimensions: task completion, spatial accuracy, detail accuracy, and visual quality, prioritizing structural layout while ignoring task-irrelevant stylistic artifacts. Finally, we evaluate our framework against a standard one-shot generation approach and a memory-less iterative baseline, BlenderAlchemy~\citep{huang2024blenderalchemy}. Both iterative methods are evaluated under the aforementioned \textit{best-of-$N$} setting~\citep{gu2025blendergym}, allowing us to analyze how the search budget $N$ impacts performance.

\paragraph{BlenderGym and SlideBench.}
For established quantitative benchmarks, we evaluate single-step 3D scene editing on BlenderGym~\citep{gu2025blendergym} and programmatic 2D document generation on SlideBench~\citep{ge2025autopresent}. We follow the standard evaluation protocols and metrics defined in their respective works to guarantee rigorous and reproducible comparisons.

\subsection{Qualitative Settings}
For complex qualitative demonstrations (\eg, full 3D scene reconstruction and 4D physics simulation), we expand the agent's memory capacity by setting \texttt{context\_window=12} and \texttt{max\_iterate\_round=100}. We utilize GPT-5~\citep{openai_gpt5_2025} as our base VLM for these long-horizon tasks. Empirically, we observe that while Claude-Sonnet-4~\citep{Anthropic2025_ClaudeSonnet4} and Gemini-2.5-Pro~\citep{GoogleDeepMind2025_Gemini2.5Pro} can complete reconstruction-from-scratch tasks with reduced visual fidelity, GPT-4o~\citep{OpenAI2024_GPT4o} struggles with active spatial exploration. Specifically, it frequently fails to adjust the camera to observe the global scene layout, rendering it unable to complete most complex reconstruction tasks. While the generational capability gap among VLMs is widely recognized, it is rarely explicitly characterized. Our tasks clearly expose this gap, translating abstract model differences into observable failures in active spatial reasoning.

\section{Extended Results}
\label{sec:appendix_extended_results}

\subsection{Extended Qualitative Results: 3D Assets and 4D Videos}
\label{sec:appendix_qualitative_supp}

To provide a more comprehensive and interactive evaluation of our framework's capabilities, we include additional qualitative results in the supplementary archive. We highly encourage reviewers to explore these files to fully assess the geometric and temporal fidelity of our generation.

\paragraph{Interactive 3D Scene Representations.}
While the main paper visualizes 2D rendered snapshots, \ours{} fundamentally operates within the inverse graphics paradigm, synthesizing real parametric 3D scenes. In the supplementary material, we provide the raw \texttt{.blend} files for a variety of reconstructed scenes. Reviewers are encouraged to open these files in the Blender software to freely navigate the camera, inspect the fine-grained geometric structures, spatial layouts, and material configurations. This interactive inspection explicitly demonstrates that the agent does not merely overfit to 2D pixel patterns, but rather grounds its reasoning to construct robust, compositional 3D assets.

\paragraph{4D Dynamic Physical Simulations.}
For the 4D physics simulation tasks, static frames cannot adequately capture temporal dynamics. Therefore, we include raw \texttt{.blend} files and rendered \texttt{.mp4} videos in the supplementary archive. These videos showcase the temporal consistency and accurate physical interactions (\eg, multi-object collisions, gravity, and glass shattering) achieved by \ours{}. By explicitly executing programmatic commands within the physics engine, our framework strictly enforces physical constraints over time, effectively bypassing the temporal flickering and physical hallucinations that plague traditional diffusion-based video generation models.

\subsection{Extended Quantitative Results}
\label{sec:appendix_quantitative_supp}

\paragraph{Comprehensive Benchmark Performance.}
We present the complete quantitative evaluation on BlenderGym (\cref{tab:supp_blendergym}) and BlenderBench (\cref{tab:supp_blenderbench}) under the best-of-1 setting. While the main paper highlights GPT-4o and Qwen3-VL-8B as representative closed- and open-source models to analyze search budgets, these tables extend the evaluation to include the full suite of foundation models: Claude-Sonnet-4 and Gemini-2.5-Pro. All experiments adhere strictly to the protocols detailed in Section 4, comparing \ours{} directly against the standard one-shot baseline.

\paragraph{Cross-Model Generalization.}
The results demonstrate that the performance gains yielded by \ours{} are highly consistent across different VLMs. On BlenderBench (\cref{tab:supp_blenderbench}), Gemini-2.5-Pro exhibits a substantial improvement in Task 2 (Fixed-View Editing), reducing the Photometric Loss from 10.14 to 1.99. Similarly, Claude-Sonnet-4 demonstrates robust error-recovery in Task 1 (Camera Adjustment), effectively rectifying the severe viewpoint misalignment that plagues its one-shot baseline. These extended results corroborate our conclusion: the interleaved multimodal reasoning loop acts as a model-agnostic capability scaffold, successfully unlocking spatial reasoning across diverse foundation models.

\begin{table}[t]
\centering
\small
\resizebox{1.0\textwidth}{!}{
    \begin{tabular}{l l cc cc cc cc cc c}
        \toprule
        \multirow{2}{*}{Model} & \multirow{2}{*}{Setting} 
        & \multicolumn{2}{c}{Blend Shape} 
        & \multicolumn{2}{c}{Placement} 
        & \multicolumn{2}{c}{Geometry} 
        & \multicolumn{2}{c}{Lighting} 
        & \multicolumn{2}{c}{Material} & \multirow{2}{*}{\makecell{Impr.\\(\%)}}\\
        \cmidrule(lr){3-4} \cmidrule(lr){5-6} \cmidrule(lr){7-8} \cmidrule(lr){9-10} \cmidrule(lr){11-12}
        & & PL $\downarrow$ & N-CLIP $\downarrow$
          & PL $\downarrow$ & N-CLIP $\downarrow$
          & PL $\downarrow$ & N-CLIP $\downarrow$
          & PL $\downarrow$ & N-CLIP $\downarrow$
          & PL $\downarrow$ & N-CLIP $\downarrow$ \\
        \midrule
        \multirow{2}{*}{GPT-4o} 
        & One-Shot & 7.94 & 19.96 & 11.86 & 29.31 & 18.12 & 24.91 & 2.06 & 2.17 & 8.78 & 15.25 & -  \\
        & \ours & \textbf{6.70} & \textbf{16.70} & \textbf{9.89} & \textbf{26.87} & \textbf{11.82} & \textbf{15.76} & \textbf{1.43} & \textbf{1.41} & \textbf{6.11} & \textbf{9.39} & 26.88 \\ 
        \midrule
        \multirow{2}{*}{Claude-Sonnet-4} 
        & One-Shot & 8.07 & 19.92 & 11.62 & 42.55 & 24.73 & 32.52 & 2.13 & 2.26 & 10.51 & 18.21 & - \\
        & \ours & \textbf{6.83} & \textbf{16.16} & \textbf{11.54} & \textbf{32.66} & \textbf{16.22} & \textbf{21.23} & \textbf{0.75} & \textbf{1.24} & \textbf{5.28} & \textbf{9.46} & 33.70 \\
        \midrule
        \multirow{2}{*}{Gemini-2.5-Pro} 
        & One-Shot & \textbf{7.01} & 17.62 & 12.39 & 37.83 & 30.73 & 39.97 & 1.95 & 2.42 & 9.28 & 15.33 & - \\
        & \ours & 7.62 & \textbf{11.73} & \textbf{10.63} & \textbf{21.11} & \textbf{9.84} & \textbf{9.71} & \textbf{0.94} & \textbf{1.15} & \textbf{2.15} & \textbf{4.96} & 47.36 \\
        \midrule
        \multirow{2}{*}{Qwen3-VL-8B} 
        & One-Shot & \textbf{7.76} & \textbf{21.78} & 14.10 & 41.44 & 11.65 & 14.02 & 12.12 & 12.30 & 9.83 & 17.16 & - \\
        & \ours & 13.51 & 27.07 & \textbf{10.07} & \textbf{28.12} & \textbf{9.15} & \textbf{12.24} & \textbf{2.33} & \textbf{2.58} & \textbf{6.43} & \textbf{11.53} & 22.64 \\
        \bottomrule
    \end{tabular}
}
\captionsetup{skip=6pt}
\caption{\textbf{Comprehensive Evaluation on BlenderGym.} We report the results of all four evaluated foundation models. Metrics include Photometric Loss (PL) and Negative-CLIP Score (N-CLIP). All the models are evaluated under the best-of-1 setting.}
\label{tab:supp_blendergym}
\end{table}

\begin{table}[t]
\centering
\small
\resizebox{\textwidth}{!}{
\begin{tabular}{l l cc cc cc cc cc c}
\toprule
\multirow{2}{*}{Model} & \multirow{2}{*}{Setting} 
& \multicolumn{3}{c}{Task 1} 
& \multicolumn{3}{c}{Task 2} 
& \multicolumn{3}{c}{Task 3} & \multirow{2}{*}{\makecell{Impr.\\(\%)}}\\
\cmidrule(lr){3-5} \cmidrule(lr){6-8} \cmidrule(lr){9-11}
& & PL $\downarrow$ & N-CLIP $\downarrow$ & VLM $\uparrow$
  & PL $\downarrow$ & N-CLIP $\downarrow$ & VLM $\uparrow$
  & PL $\downarrow$ & N-CLIP $\downarrow$ & VLM $\uparrow$ \\
\midrule
\multirow{2}{*}{GPT-4o} 
& One-Shot & 48.16 & 64.17 & 0.58 & 7.36 & 7.12 & 2.75 & 30.14 & 38.69 & 0.25 & - \\
& \ours & \textbf{8.56} & \textbf{18.19} & \textbf{1.44} & \textbf{5.11} & \textbf{4.33} & \textbf{3.58} & \textbf{14.51} & \textbf{17.91} & \textbf{1.53} & 113.96 \\
\midrule
\multirow{2}{*}{Claude-Sonnet-4} 
& One-Shot & 20.26 & 33.82 & 1.36 & 4.47 & 3.16 & 2.75 & 26.66 & 32.80 & 1.44 & - \\
& \ours & \textbf{9.42} & \textbf{11.62} & \textbf{2.47} & \textbf{1.34} & \textbf{2.51} & \textbf{3.67} & \textbf{8.98} & \textbf{11.97} & \textbf{1.80} & 53.07 \\
\midrule
\multirow{2}{*}{Gemini-2.5-Pro} 
& One-Shot & 40.65 & 49.57 & 1.75 & 10.14 & 13.00 & 2.78 & 31.66 & 39.89 & 0.64 & - \\
& \ours & \textbf{17.77} & \textbf{20.58} & \textbf{2.33} & \textbf{1.99} & \textbf{3.64} & \textbf{3.75} & \textbf{27.41} & \textbf{30.97} & \textbf{0.66} & 41.12 \\
\midrule
\multirow{2}{*}{Qwen3-VL-8B} 
& One-Shot & 60.82 & 78.16 & 0.28 & 33.14 & 37.51 & 1.61 & 22.81 & 22.98 & 1.25 & - \\
& \ours & \textbf{10.54} & \textbf{17.27} & \textbf{1.31} & \textbf{3.85} & \textbf{3.34} & \textbf{3.33} & \textbf{11.35} & \textbf{6.86} & \textbf{2.25} & 112.79 \\
\bottomrule
\end{tabular}
}
\captionsetup{skip=6pt}
\caption{\textbf{Comprehensive Evaluation on BlenderBench.} We report the results of all four evaluated foundation models. Metrics include Photometric Loss (PL), Negative-CLIP Score (N-CLIP), and VLM Score (VLM). All the models are evaluated under the best-of-1 setting.}
\label{tab:supp_blenderbench}
\end{table}

\section{System Prompts}
\label{sec:appendix_prompts}

Previous VLM-based procedural generation methods~\citep{ge2025autopresent, ling2025scenethesis, sun20253dgeneralistselfimprovingvisionlanguageactionmodels} heavily rely on complex, task-specific prompt engineering to inject domain priors. In contrast, \ours{} utilizes a concise, unified prompt structure across all evaluated tasks, modifying only a few domain-specific keywords (\eg, 2D $\rightarrow$ 3D) to adapt to different environments. 
During the interleaved reasoning loop, the agent seamlessly transitions between modifying the state and observing the results. To facilitate this, we provide the agent with focused system instructions depending on its current active phase:

\paragraph{Instructions for Code Synthesis (State Modification).}
When the agent is tasked with synthesizing or editing the scene program, it receives the following base instruction:

\begin{promptbox}
[Role]

You are an expert, tool-driven coding agent that builds 3D static scenes. You will receive (a) an image describing the target scene and (b) an optional text description. Your goal is to reproduce the target scene as faithfully as possible.

[Response Format]

The task proceeds over multiple rounds. In each round, your response must be exactly one tool call with reasoning in the content field. If you would like to call multiple tools, you can call them one by one in the following turns. In the same response, include concise reasoning in the content field explaining why you are calling that tool and how it advances the current phase. Always return both the tool call and the content together in one response.
\end{promptbox}

\paragraph{Instructions for Visual Inspection (State Observation).}
Upon rendering the scene, the agent transitions into the inspection phase to actively gather visual evidence. It receives the following instruction:

\begin{promptbox}
[Role]

You are an expert visual inspector of 3D scenes. You will receive:
(1) A description of the target scene, including (a) an image describing the target scene and (b) an optional text description.
In each following round, you will receive the current scene context, including (a) the overall structural plan, (b) the step-by-step code edits executed so far (including your previous thoughts and the full code), and (c) the current scene render(s) produced by the engine.
Your task is to use read-only perceptual tools to precisely and comprehensively analyze discrepancies between the current scene and the target, and to propose actionable next-step modifications for the code editing phase.

[Response Format]

The task proceeds over multiple rounds. In each round, your response must be exactly one tool call with reasoning in the content field. If you would like to call multiple tools, you can call them one by one in the following turns. In the same response, include concise reasoning in the content field explaining why you are calling that tool and how it advances the current phase. Always return both the tool call and the content together in one response.
\end{promptbox}

\paragraph{VLM Score Evaluator Prompt.}
In BlenderBench, the \textit{VLM Score} is computed using GPT-4o as an impartial evaluator. The evaluation is dynamically grounded by injecting specific parameters into the prompt at runtime: 
\begin{itemize}
    \item \texttt{\{criteria\}}: The specific fine-grained dimension being evaluated (\ie, task completion, visual quality, spatial accuracy, or detail accuracy) along with its detailed scoring rubric.
    \item \texttt{\{task\_description\}}: The exact text instruction provided for the specific test instance.
    \item \texttt{\{scale\}}: The maximum score bounds for the integer rating (typically set to 5).
\end{itemize}

The complete evaluation prompt template is formulated as follows:

\begin{promptbox}
You are evaluating a 3D scene image generated based on a specific task description. You will be shown a TARGET image (may be style-transferred/noisy) and a GENERATED image. Compare them with respect to the task, prioritizing geometric structure, spatial layout, object identity/pose/placement, and lighting intent over stylistic differences (colors, textures, style effects).

\{criteria\}

Task Description: \{task\_description\}

Instructions:
- Use the TARGET image as a visual reference for intended scene layout and camera/view configuration.
- Ignore style-transfer artifacts (e.g., color grading, texture stylization, artistic filters) when they do not affect task objectives.
- Focus your judgment on whether the GENERATED image meets the task criteria relative to the TARGET reference.

Give an integer score between 0 - \{scale\}, where higher scores mean the criteria is better met.
First, respond with a score; Then, provide your justification for the score in natural language sentences. Your response should look like this: ``4. The cabinet is correctly positioned and the plant is visible in the mirror as required.''
Only evaluate the image based on the specified criteria, and no other aspects. Give scores across the full spectrum (0-5) instead of only good ones (3-5).
\end{promptbox}

\section{Semantic Skill Library Schemas}
\label{sec:appendix_skills}

As detailed in the main text, \ours{} leverages a high-level semantic skill library rather than raw, low-level graphics APIs. This abstraction significantly compresses the action space and insulates the agent from execution trivialities. Here, we provide the complete JSON Schemas for these encapsulated interfaces. To seamlessly support the interleaved reasoning loop, the library is explicitly bifurcated into two categories based on the agent's active phase:

\paragraph{State Modification Interfaces.}
During the code synthesis phase, the agent is equipped with the following execution-oriented skills to instantiate assets and apply programmatic edits to the scene:

\begin{lstlisting}[language=json]
"name": "execute_code",
"description": "Execute blender python code and trigger verifier evaluation.\nReturns either:\n(1) On error: detailed error information; or \n(2) On success: a clear render (you must add a camera in your code) and further modification suggestions from a separate verifier agent.",
"parameters": {
    "thought": {
        "type": "string",
        "description": "Think step by step about the current scene and reason about what code to write next. Describe your reasoning process clearly."
    },
    "code_diff": {
        "type": "string",
        "description": "Before outputting the final code, precisely list the line-level edits you will make. Use this minimal diff-like format ONLY:\n\n-: [lines to remove]\n+: [lines to add]\n\nRules:\n1) Show only the smallest necessary edits (avoid unrelated changes).\n2) Keep ordering: list removals first, then additions.\n3) Do not include commentary here-only the edit blocks.\n4) If starting from scratch, use `-: []` and put all new lines under `+: [...]`.\n5) Every line is a literal code line (no markdown, no fences)."
    },
    "code": {
        "type": "string",
        "description": "Provide the COMPLETE, UPDATED Blender Python code AFTER applying the edits listed in `code_diff`. The full code must include both the modified lines and the unchanged lines to ensure a coherent, runnable script."
    }
}
"required": ["thought", "code_diff", "code"]
\end{lstlisting}

\begin{lstlisting}[language=json]
"name": "get_scene_info",
"description": "Get the scene information including objects, materials, lights, and cameras. This tool provides detailed information about the current state of the Blender scene, which can be used to understand what objects exist and their properties.",
"parameters": {}
\end{lstlisting}

\begin{lstlisting}[language=json]
"name": "end_process",
"description": "No-op tool used to indicate the process should end. If the scene has no remaining issues, stop making changes and call this tool.",
"parameters": {}
\end{lstlisting}

\begin{lstlisting}[language=json]
"name": "make_plan",
"description": "From the given inputs, imagine and articulate the scene in detail. This tool does not return new information. It stores your detailed description as your own plan to guide subsequent actions. You must call this tool first.",
"parameters": {
    "overall_description": {
        "type": "string", 
        "description": "A thorough, comprehensive depiction of the entire scene.\nExample (Simple Room - Overall Description): \"A compact, modern study room measuring 4.0 m (X) x 3.0 m (Y) x 2.8 m (Z), with the world origin at the center of the floor. Walls are matte white (slightly warm); the floor is light-gray concrete with subtle roughness; the ceiling is white. The +Y side is the 'north wall', -Y is 'south', +X is 'east', -X is 'west'. A single rectangular window (1.2 m x 1.0 m) is centered on the west wall (X = -2.0 m plane), sill height 0.9 m from the floor, with a thin black metal frame and frosted glass that softly diffuses daylight. Primary furniture: a medium-tone oak desk against the north wall, a simple black task chair, a slim floor lamp to the desk's right, and a low potted plant softening the corner. A framed A2 poster hangs above the desk, and a 1.6 m x 1.0 m flat-woven rug (light beige) sits beneath the desk area. Lighting combines soft daylight from the window with a warm key from the floor lamp; the ambience is calm, minimal, and functional.\""
    },
    "detailed_plan": {
        "type": "string", 
        "description": "Consider a detailed plan for scene construction. This plan should follow this format:\n1. Preparation Stage: Use the appropriate tool to generate and download the necessary 3D assets, which are typically complex objects that cannot be constructed using basic geometry.\n2. Rough Stage: Establish the global layout and basic environment components, including the floor, walls or background, camera, and main light source.\n3. Intermediate Stage: Import the downloaded objects into the scene, adjusting their positions, scales, and orientations to align with the global layout. Construct any missing objects using basic geometry.\n4. Refinement Stage: Refine details, enhance materials, add auxiliary lights and props, and make precise local adjustments to enhance realism and accuracy."
    }
},
"required": ["overall_description", "detailed_plan"]
\end{lstlisting}

\begin{lstlisting}[language=json]
"name": "get_better_object",
"description": "Generate high-quality 3D assets, download them locally, and provide their paths for later use. The textures, materials, and finishes of these objects are already high-quality with fine-grained detail; please do not repaint them. If you do, you will need to re-import the object.",
"parameters": {
    "thought": {
        "type": "string", 
        "description": "Think about the object you want to download. Consider the overall description of the scene and the detailed plan for scene construction."
    },
    "object_name": {
        "type": "string", 
        "description": "The name of the object to download. For example, 'chair', 'table', 'lamp', etc."
    },
    "reference_type": {
        "type": "string", 
        "enum": ["text", "image"], 
        "description": 'The type of generation reference. If the target 3D asset in the reference image is clear and unobstructed, use reference_type=\"image\". Otherwise, use reference_type=\"text\".'
    },
    "object_description": {
        "type": "string", 
        "description": "If you use reference_type=\"text\", you must provide a detailed description of the object to download."
    },
    "rig_and_animate": {
        "type": "boolean", 
        "description": "Whether to rig and animate the downloaded asset. True for dynamic scene, False for static scene"
    },
    "action_description": {
        "type": "string", 
        "description": "If you use rig_and_animate=True, you must provide a description of the action to apply to the downloaded asset. Only input verbs here, e.g. walk, run, jump, etc."
    }
},
"required": ["thought", "object_name"]
\end{lstlisting}

\paragraph{State Observation Interfaces.}
During the visual inspection phase, the agent relies on the following read-only, information-seeking skills. These allow the agent to actively adjust viewpoints, probe the rendered scene, and gather fine-grained visual evidence without altering the underlying geometric state:

\begin{lstlisting}[language=json]
"name": "initialize_viewpoint",
"description": "Adds a viewpoint to observe the listed objects. The viewpoints are added to the four corners of the bounding box of the listed objects. This tool returns the positions and rotations of the four viewpoint cameras, as well as the rendered images of the four cameras. You can use these information to set the camera to a good initial position and orientation.",
"parameters": {
    "object_names": {
        "type": "array", 
        "description": "The names of the objects to observe. Objects must exist in the scene (you can check the scene information to see if they exist). If you want to observe the whole scene, you can pass an empty list.",
        "items": {
            "type": "string",
            "description": "The name of the object to observe."
        }
    }
},
"required": ["object_names"]
\end{lstlisting}

\begin{lstlisting}[language=json]
"name": "set_camera",
"description": "Set the current active camera to the given location and rotation",
"parameters": {
    "location": {
        "type": "array", 
        "description": "The location of the camera (in world coordinates)",
        "items": {
            "type": "number",
            "description": "The location of the camera (in world coordinates)"
        }
    },
    "rotation_euler": {
        "type": "array", 
        "description": "The rotation of the camera (in euler angles)",
        "items": {
            "type": "number",
            "description": "The rotation of the camera (in euler angles)"
        }
    }
},
"required": ["location", "rotation_euler"]
\end{lstlisting}

\begin{lstlisting}[language=json]
"name": "investigate",
"description": "Investigate the scene by the current camera. You can zoom, move, and focus on the object you want to investigate.",
"parameters": {
    "operation": {
        "type": "string", 
        "enum": ["zoom", "move", "focus"], 
        "description": "The operation to perform."},
    "object_name": {
        "type": "string", 
        "description": "If the operation is focus, you need to provide the name of the object to focus on. The object must exist in the scene."
    },
    "direction": {
        "type": "string", 
        "enum": ["up", "down", "left", "right", "in", "out"], 
        "description": "If the operation is move or zoom, you need to provide the direction to move or zoom."
    }
},
"required": ["operation"]
\end{lstlisting}

\begin{lstlisting}[language=json]
"name": "set_visibility",
"description": "Set the visibility of the objects in the scene. You can decide to show or hide the objects. You do not need to mention all the objects here, the objects you do not metioned will keep their original visibility.",
"parameters": {
    "show_objects": {
        "type": "array", 
        "description": "The names of the objects to show. Objects must exist in the scene.",
        "items": {
            "type": "string",
            "description": "The name of the object to show."
        }
    },
    "hide_objects": {
        "type": "array", 
        "description": "The names of the objects to hide. Objects must exist in the scene.",
        "items": {
            "type": "string",
            "description": "The name of the object to hide."
        }
    }
},
"required": ["show_objects", "hide_objects"]
\end{lstlisting}

\begin{lstlisting}[language=json]
"name": "set_keyframe",
"description": "Set the scene to a specific frame number for observation",
"parameters": {
    "frame_number": {
        "type": "integer", 
        "description": "The specific frame number to set the scene to."
    }
},
"required": ["frame_number"]
\end{lstlisting}

\begin{lstlisting}[language=json]
"name": "get_scene_info",
"description": "Get the scene information",
"parameters": {}
\end{lstlisting}

\begin{lstlisting}[language=json]
"name": "end_process",
"description": "If you think your observations are sufficient, call the tool to end the process and output the answer.",
"parameters": {
    "visual_difference": {
        "type": "string", 
        "description": "Visual difference between the current scene and the target scene. You can answer from the following four aspects:\n1) Camera\n- If the generator's camera choice is poor (occlusions, missing key objects, suboptimal angle), use setup_camera and investigate to find a better viewpoint.\n- Provide the exact observer camera coordinates and orientation as part of your edit_suggestion so the generator can replicate or adapt them.\n2) Objects\n- Verify that all key objects in the target image exist in the current scene.\n- If objects are missing or extraneous, recommend additions or removals and specify whether to generate a new asset or duplicate an existing one.\n- If a present object diverges materially from the target (e.g., target chair is black but current chair is white), recommend replacement or material edits as appropriate.\n3) Layout\n- Check whether spatial layout matches the target.\n- If not, recommend concrete transforms (move/rotate/scale) and, when possible, indicate relative or absolute adjustments (e.g., "move desk 0.3 m toward +Y," "rotate lamp -15 degrees around Z").\n4) Environment\n- Assess background, lighting direction/intensity, and overall ambience.\n- If these do not match the target, recommend changes (e.g., environment map, key/fill/rim balance, wall/floor backdrop corrections).\n5) Animation (only required in dynamic mode)\n- Assess whether animated objects exhibit correct motion; verify inter-object interactions (e.g., contact timing, grasp/kick/hold), per-frame physical plausibility, and whether the overall keyframe sequence satisfies the intended action.\n- Use set_key_frame to inspect critical frames; reference exact frame indices from the scene metadata in your suggestions."
    },
    "edit_suggestion": {
        "type": "string", 
        "description": "Edit suggestion for the current scene. Refer to the visual difference to propose the edit suggestion."
    }
},
"required": ["visual_difference", "edit_suggestion"]
\end{lstlisting}

\end{document}